\documentclass[10pt,twocolumn,letterpaper]{article}

\usepackage[pagenumbers]{cvpr}      

\usepackage{cuted}
\usepackage{amsthm}

\newtheorem{proposition}{Proposition}

\definecolor{cvprblue}{rgb}{0.21,0.49,0.74}
\usepackage[pagebackref,breaklinks,colorlinks,allcolors=cvprblue]{hyperref}

\def\confName{CVPR}
\def\confYear{2027}

\title{Borrow the Style, Follow the Prompt: Progressive Reference-Guided Stylization}

\author{Yu Lin \quad Zi-Jun Wang \quad Zhilin Zhao\\
	Sun Yat-sen University\\
}

\begin{document}
\maketitle
\begin{strip}
\centering

\includegraphics[width=\linewidth]{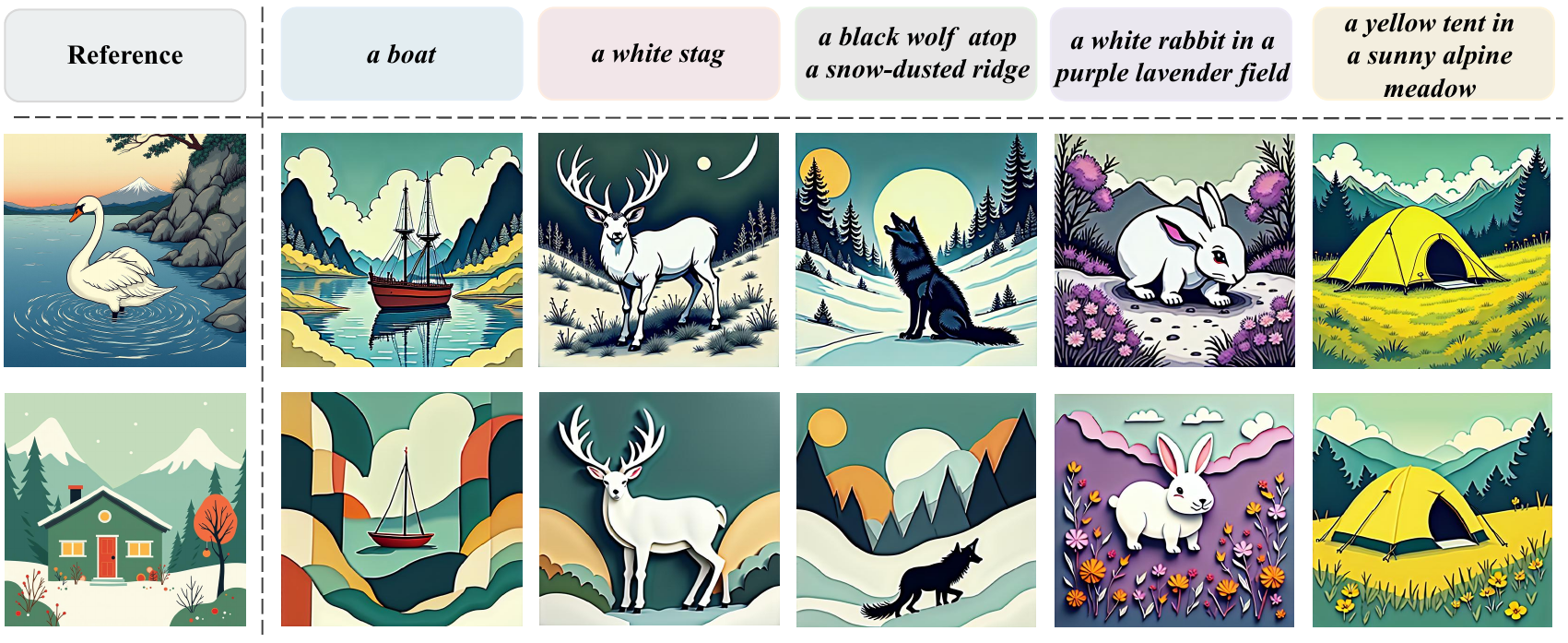}

\captionsetup{hypcap=false}
\captionof{figure}{
Given a reference image and diverse target prompts,
ProStyle transfers the reference style across different target contents
while faithfully following the target prompts.
}
\label{fig:teaser}
\captionsetup{hypcap=true}

\end{strip}
\begin{abstract}
Reference-guided text-to-image stylization transfers visual appearance to content specified by text, but the reference subjects and backgrounds can interfere with what the generated image depicts. Suppressing this unwanted information can also discard transferable texture and material cues, creating a tension between content control and faithful preservation of the reference style. Accordingly, we propose \textbf{ProStyle}, which learns a content-suppressed style prior through cross-content supervision on same-style pairs with different subjects and backgrounds and injects it with weak conditioning at auxiliary layers to strengthen style expression. Building on this prior, a complementary pathway learns residual corrections from full reference patch features with the content-suppressed style pathway frozen, before both pathways undergo joint refinement under the same objective. For cross-content learning and evaluation, we introduce \textbf{ReConStyle}, which provides 134K training pairs, two test settings, and object-level measures of target-content fidelity and reference-object overlap alongside style similarity and text alignment. ProStyle outperforms all compared SD1.5-based baselines and achieves the best average rank of 1.86 across four automatic and three human-evaluation criteria on ReConStyle-Full. In the user study, it substantially outperforms the baselines in text alignment and combined suitability, with selection rates of 84.36\% and 50.90\%, respectively. On the external Style-Rank benchmark, it also achieves the highest CLIP-Text and ImageReward scores.
\end{abstract}
    
\section{Introduction}
\label{sec:introduction}

Reference-guided stylization combines the content control of text-to-image generation with appearance cues supplied by an image~\cite{rombach2022high,saharia2022photorealistic,ye2023ip}. For example, a papercraft guitar reference should lend layered texture to a red fox requested by text without carrying the guitar into the generated scene. Achieving this transfer is difficult because the reference encodes texture, material, and shape treatment together with the objects and background that exhibit them.
\begin{figure}[t]
    \centering
    \includegraphics[width=\linewidth]{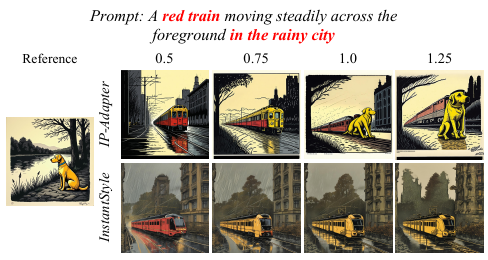}
    \caption{
    Effect of reference-conditioning strength $\alpha$ on IP-Adapter and InstantStyle. Stronger conditioning introduces the reference dog in IP-Adapter and shifts the requested red train toward yellow in InstantStyle. Quantitative results are in Figure~\ref{fig:strength_curves}.
    }
    \label{fig:intro_examples}
\end{figure}

Existing methods regulate this mixed information through visual adapters, attention, compact style representations, or selective injection~\cite{ye2023ip,hertz2024style,jeong2024visual,qi2024deadiff,wang2024instantstyle}. At the training level, DEADiff learns style representations from same-prompt image pairs~\cite{qi2024deadiff}, whereas CSGO relies on content--style--stylized triplets to separate content and style conditioning~\cite{xing2024csgo}. UniCSG further combines staged semantic disentanglement with reward learning~\cite{yang2026unicsg}. Masking, feature purification, and residual control further suppress unwanted reference semantics~\cite{zhu2025less,feng2026cleanstyle,yang2026safestyle}, as discussed in Appendix~\ref{sec:related_work}. However, suppressing reference-specific content may also remove transferable stylistic details. Retaining richer reference features makes more appearance information available while creating additional routes for reference objects to enter the output. Figure~\ref{fig:intro_examples} illustrates this tension, showing how stronger reference conditioning can introduce reference objects or override text-specified attributes. The remaining challenge is therefore to integrate useful reference information while retaining control over the requested content.

We address this tension with \textbf{ProStyle}, which first learns a content-suppressed style prior and then progressively integrates richer reference features through a complementary residual pathway. When a reference and its training target share subjects or backgrounds, the model may reuse this content to fit the target. We therefore pair same-style images with different subjects and backgrounds, so that reference content offers little help in fitting the target. Hierarchical asymmetric reference injection applies the learned style tokens strongly at primary layers and weakly at auxiliary layers, extending style conditioning to all cross-attention layers while limiting interference with the requested content. Because a compact style representation may lose appearance details, a complementary native reference pathway learns residual corrections from full reference patch features while the style pathway is frozen, followed by joint refinement of both pathways under the same cross-content noise-prediction objective. A denoising-risk analysis characterizes reference-content sensitivity under conditional mean independence and the best attainable risk with complementary reference inputs.

To provide such pairs and assess the resulting trade-off, we construct \textbf{ReConStyle}, which combines 134K cross-content training pairs with two test settings and object-level measures of target-content fidelity and reference-object overlap alongside style similarity and text alignment. On ReConStyle-Full, ProStyle outperforms all same-backbone SD1.5 baselines (DEADiff, StyleShot, and Less is More) on all four automatic metrics, remains close to the best style and content scores of methods built on stronger SDXL and SD3.5-Large backbones, and achieves the best average rank across the seven evaluation criteria. In the user study, ProStyle exceeds the next-best methods in text-alignment and combined-suitability selection rates by 30.64 and 36.03 percentage points, respectively. On the external Style-Rank benchmark~\cite{benaroche2024stylerank}, ProStyle also achieves the highest CLIP-Text~\cite{radford2021learning} and ImageReward~\cite{xu2023imagereward} scores. Our contributions are as follows:
\begin{enumerate}
    \item We construct ReConStyle, a benchmark with 134K cross-content training pairs that share a style but differ in subject and background, two test settings, and object-level measures of target-content fidelity and reference-object overlap.
    \item We develop ProStyle, a progressive framework that learns a content-suppressed style prior, fits full-reference residuals with this prior fixed, and then jointly refines both pathways.
    \item We introduce weak auxiliary-layer conditioning, which injects the learned style tokens into auxiliary layers at a small scale to strengthen style expression while preserving the requested content.
\end{enumerate}

\section{ReConStyle Benchmark}
\label{sec:reconstyle_benchmark}

\begin{figure*}[t]
    \centering
    \includegraphics[width=\textwidth]{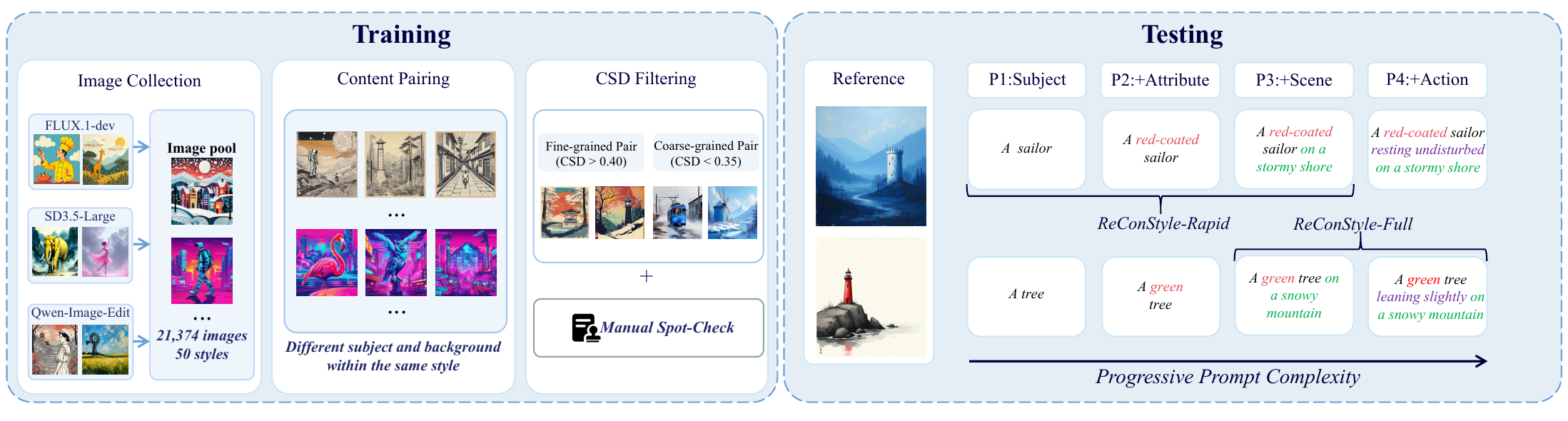}
    \caption{
    Overview of ReConStyle construction and evaluation. Training shows image collection, cross-content pairing, and CSD filtering with manual spot-checks. Testing illustrates prompt levels. P4 adds action and spatial placement, and its examples show the action component.
    }
    \label{fig:dataset_benchmark}
\end{figure*}

Learning a content-suppressed style prior requires training pairs that share only style, and evaluating it requires joint assessment of style similarity and content control. Existing evaluations assess style consistency and text alignment, and recent work also quantifies reference-content leakage~\cite{qi2024deadiff,lei2025stylestudio,aravanis2025only}. However, few benchmarks jointly evaluate target-subject fidelity and reference-object overlap at the object level, and existing paired data mainly target content-image-conditioned stylization~\cite{xing2024csgo}. We introduce \textbf{ReConStyle}, combining cross-content training pairs, two test settings, and complementary style and content metrics.

\subsection{Training Set}
\label{sec:training_set}

Each training pair consists of two images from the same style category that differ in both subject and background. The training collection contains 21,374 generated images spanning 50 visual styles. It includes 16,000 images generated evenly by FLUX.1-dev~\cite{blackforestlabs2024flux} and Stable Diffusion 3.5 Large~\cite{stabilityai2024sd35}, together with 5,374 supplementary Qwen-Image-Edit 2511~\cite{wu2025qwenimage} outputs.

To capture different levels of intra-style similarity, we divide candidate pairs using CSD~\cite{somepalli2024investigating}. Pairs with CSD $>0.40$ are treated as fine-grained, and those with CSD $<0.35$ as coarse-grained. Manual spot-checking supports the stylistic compatibility of sampled pairs with different contents. This process yields 93,800 fine-grained and 40,200 coarse-grained pairs, for a total of 134,000 cross-content training pairs. Appendix~\ref{sec:supp_benchmark} details the full construction procedure.

\subsection{Test Set}
\label{sec:test_set}

\paragraph{ReConStyle-Rapid.}
This setting supports fast validation during development and ablation. It contains 100 synthetic references from 10 styles. For each reference, three progressively specified prompts P1--P3 share the same target subject, yielding 300 conditions.

\paragraph{ReConStyle-Full.}
This setting supports large-scale evaluation and covers 50 visual styles with 3,300 reference images, comprising 2,800 synthetic images and 500 manually selected images from Style30K~\cite{li2024styletokenizer}. Each style contains 66 references, and each reference is paired with two independently constructed structured prompts. P3 specifies a target subject, subject attribute, scene attribute, and background, while P4 additionally incorporates action and spatial placement constraints, yielding 6,600 evaluation conditions. Full includes 40 seen styles and 10 unseen styles. All evaluation prompts are checked against the training prompt pool to avoid overlapping content descriptions and normalized combinations of subject and background. Because synthetic references are generated from known content descriptions, their subjects are also required to differ from the target subjects.

\subsection{Evaluation Metrics}
\label{sec:evaluation_metrics}

Style-CSD measures CSD similarity to subject-free exemplars of the reference style category~\cite{somepalli2024investigating}. Comparing with category exemplars reduces the influence of reference subjects on the score, and the user study in Section~\ref{sec:user_study} additionally judges style consistency against the original reference. CLIP-T measures cosine similarity between generated images and target prompts~\cite{radford2021learning}, and the same study complements it with human judgments of text alignment against the prompt.

Because CLIP-T scores the entire prompt, it may be less sensitive to fine-grained mismatches in subjects, attributes, and objects~\cite{ghosh2023geneval}. We therefore introduce two object-level metrics based on Qwen3-VL~\cite{bai2025qwen3}. The VLM extracts object categories and attributes from each image alone, excluding broad regions such as sky, water surfaces, fields, and roads. TCF assesses the target subject and its appearance, and ROO measures non-target reference-object overlap.

\paragraph{Target-Content Fidelity (TCF).}
Let $M_s\in\{0,1\}$ indicate the presence of the annotated target subject, and $M_{\mathrm{attr}}\in\{0,0.5,1\}$ the degree of attribute matching. TCF is defined as
\begin{equation}
\mathrm{TCF}=M_s(0.5+0.5M_{\mathrm{attr}}).
\end{equation}
TCF provides an explicit check of subject presence and appearance attributes, while CLIP-T supplies a complementary measure of overall alignment with the full prompt, including its scene, action, and spatial descriptions.

\begin{figure*}[t]
    \centering
    \includegraphics[width=\textwidth]{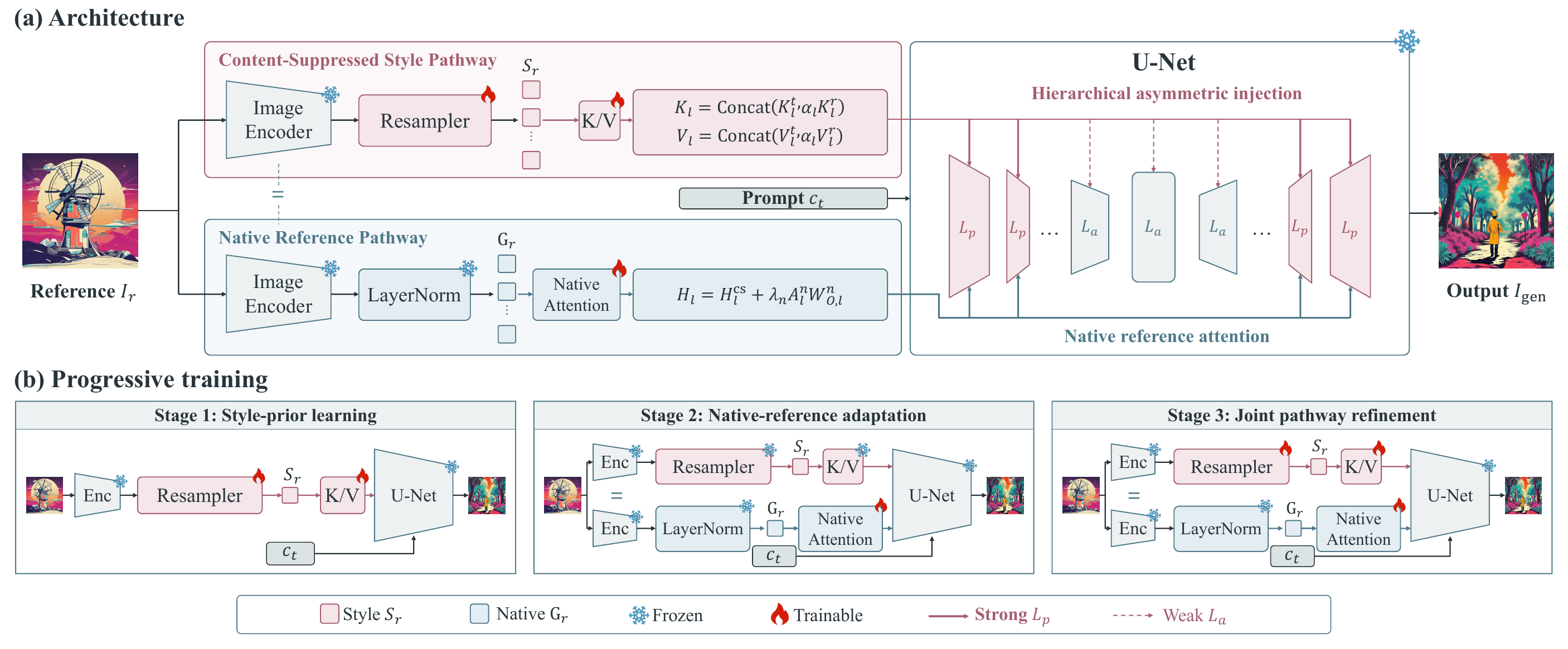}
    \caption{
    Overview of \textbf{ProStyle}.
    (a) The content-suppressed style pathway is injected strongly at primary layers and weakly at auxiliary layers. The native reference pathway provides richer reference information.
    (b) Progressive training through prior learning, residual fitting with the style pathway fixed, and joint refinement.
    }
    \label{fig:method_overview}
\end{figure*}
\paragraph{Reference-Object Overlap (ROO).}
Let $\mathcal{R}$ and $\mathcal{G}$ denote the object-category sets extracted from the reference and generated images, and $s_t$ the target subject. After the category matching in Appendix~\ref{sec:supp_roo}, ROO measures the proportion of non-target reference objects that appear in the generated image as
\begin{equation}
\mathrm{ROO}=
\frac{|\mathcal{G}\cap(\mathcal{R}\setminus\{s_t\})|}
{|\mathcal{R}\setminus\{s_t\}|}.
\end{equation}
ROO serves as an object-level indicator of reference-content leakage. References without non-target objects are excluded from the ROO average.

Higher TCF and lower ROO are better, and fixed rules compute both scores from the VLM extractions. We further validate these object-level metrics against human annotations. On 270 randomly sampled evaluation conditions across four methods, agreement between human-derived and VLM-derived scores, averaged across three annotators, reaches 79.1\% for TCF and 87.3\% for ROO. Across eight methods, TCF rankings also correlate strongly with human text-alignment rankings (Spearman $\rho=0.90$). Appendix~\ref{sec:supp_vlm_reliability} provides the complete validation protocol.

\section{Progressive Style Learning}
\label{sec:method}

Given a reference image and a target content prompt, ProStyle conditions a frozen diffusion model to synthesize the requested content in the reference style. As summarized in Figure~\ref{fig:method_overview}, a content-suppressed style pathway (\emph{style pathway} for short) learns a compact style prior through cross-content supervision and conditions the backbone through hierarchical asymmetric reference injection. A complementary native reference pathway (\emph{native pathway} for short) supplies residual corrections from all reference patches, and both pathways are trained progressively. Section~\ref{sec:prostyle_theory_main} analyzes this design.

\subsection{Cross-Content Style Prior Learning}
\label{sec:cross_content_style_prior_learning}

The style pathway aims to extract transferable style features while reducing dependence on reference-specific content. Its core mechanism is cross-content supervision. The reference and target images share the same style but differ in subject and background. Reusing the reference subject or background therefore offers little help in fitting the target.

Our training samples consist of a reference image, a target image, and a content-only target description, forming a tuple $(I_r,I_t,c_t)\sim\mathcal{D}_{\mathrm{CC}}$. Pairs follow the protocol in Section~\ref{sec:training_set}. A frozen CLIP image encoder extracts patch-level features from the reference image, excluding the class token, as
\begin{equation}
F_r=E_{\mathrm{CLIP}}(I_r)_{\mathrm{patch}}.
\label{eq:clip_reference}
\end{equation}
A trainable Resampler~\cite{alayrac2022flamingo,ye2023ip} then compresses these features into 16 style tokens
\begin{equation}
S_r=R_\phi(F_r).
\label{eq:style_tokens}
\end{equation}

Cross-content supervision discourages dependence on reference-specific content, while the compact token bottleneck limits representational capacity.

\subsection{Hierarchical Asymmetric Reference Injection}
\label{sec:hierarchical_asymmetric_reference_injection}

To propagate style more broadly through the denoiser, the learned tokens condition all cross-attention layers. Prior studies show that these layers play different roles in style and semantic control~\cite{wang2024instantstyle,qi2024deadiff}. Based on this distinction, we apply strong reference conditioning at primary layers and weak conditioning at the remaining auxiliary layers.

At layer $l$, the reference keys and values $K_l^r$ and $V_l^r$ projected from $S_r$ are concatenated with the text keys and values along the token dimension, and the attention output is computed by
\begin{equation}
\begin{aligned}
K_l&=\operatorname{Concat}(K_l^t,\alpha_l K_l^r),\\
V_l&=\operatorname{Concat}(V_l^t,\alpha_l V_l^r),\\
A_l^{\mathrm{cs}}&=
\operatorname{Softmax}\!\left(\frac{Q_lK_l^\top}{\sqrt{d_l}}\right)V_l.
\end{aligned}
\label{eq:asymmetric_injection}
\end{equation}
We designate 11 of the 16 cross-attention layers as primary layers $\mathcal{L}_p$ and the remaining five as auxiliary layers $\mathcal{L}_a$, and set
\begin{equation}
\alpha_l=
\begin{cases}
1.0,&l\in\mathcal{L}_p,\\
0.05,&l\in\mathcal{L}_a.
\end{cases}
\label{eq:layer_weight}
\end{equation}
The exact indices are provided in Appendix~\ref{sec:supp_prostyle_training}, and Appendices~\ref{sec:supp_auxiliary_ablation}--\ref{sec:supp_native_weight} study the auxiliary scale, layer selection, token count, and native pathway weight.

\subsection{Complementary Native Reference Pathway}
\label{sec:prostyle_native_pathway}

A compact style representation may lose some style details while suppressing reference content. To supplement the style-token conditioning, the native pathway exposes the primary layers to all reference patches. It uses its own reference projections and attention normalization, and contributes a residual update in the same feature space as the style pathway.

We normalize the original patch features without reducing their token count as
\begin{equation}
G_r=\operatorname{LN}(F_r).
\label{eq:prostyle_native_features}
\end{equation}
At each primary layer $l\in\mathcal{L}_p$, the native pathway obtains keys and values through independent projections and shares the frozen query projection of the pretrained attention layer, giving
\begin{equation}
\begin{aligned}
K_l^n&=G_rW_{K,l}^n,\qquad V_l^n=G_rW_{V,l}^n,\\
A_l^n&=
\operatorname{Softmax}\!\left(\frac{Q_l(K_l^n)^\top}{\sqrt{d_l}}\right)V_l^n.
\end{aligned}
\label{eq:native_attention}
\end{equation}
The outputs of the two pathways are combined as
\begin{equation}
H_l=
\begin{cases}
H_l^{\mathrm{cs}}+\lambda_n\overline{A}_l^nW_{O,l}^n,
    &l\in\mathcal{L}_p,\\
H_l^{\mathrm{cs}},&l\in\mathcal{L}_a,
\end{cases}
\label{eq:dual_pathway}
\end{equation}
where $H_l^{\mathrm{cs}}$ is the projected output of $A_l^{\mathrm{cs}}$ and $\overline{A}_l^n$ concatenates the native attention heads. We set $\lambda_n=1$ and zero-initialize the native output projection $W_{O,l}^n$, as in ControlNet~\cite{zhang2023adding}, so the native pathway initially preserves the complete style-pathway predictor.

The native pathway uses the same cross-content objective, so its residual corrections are supervised by the target image, which differs from the reference in subject and background. Both pathways share a single frozen image-encoder forward pass.

\subsection{Progressive Optimization}
\label{sec:progressive_reference_learning}

\begin{figure*}[t]
    \centering
    \includegraphics[width=\textwidth]{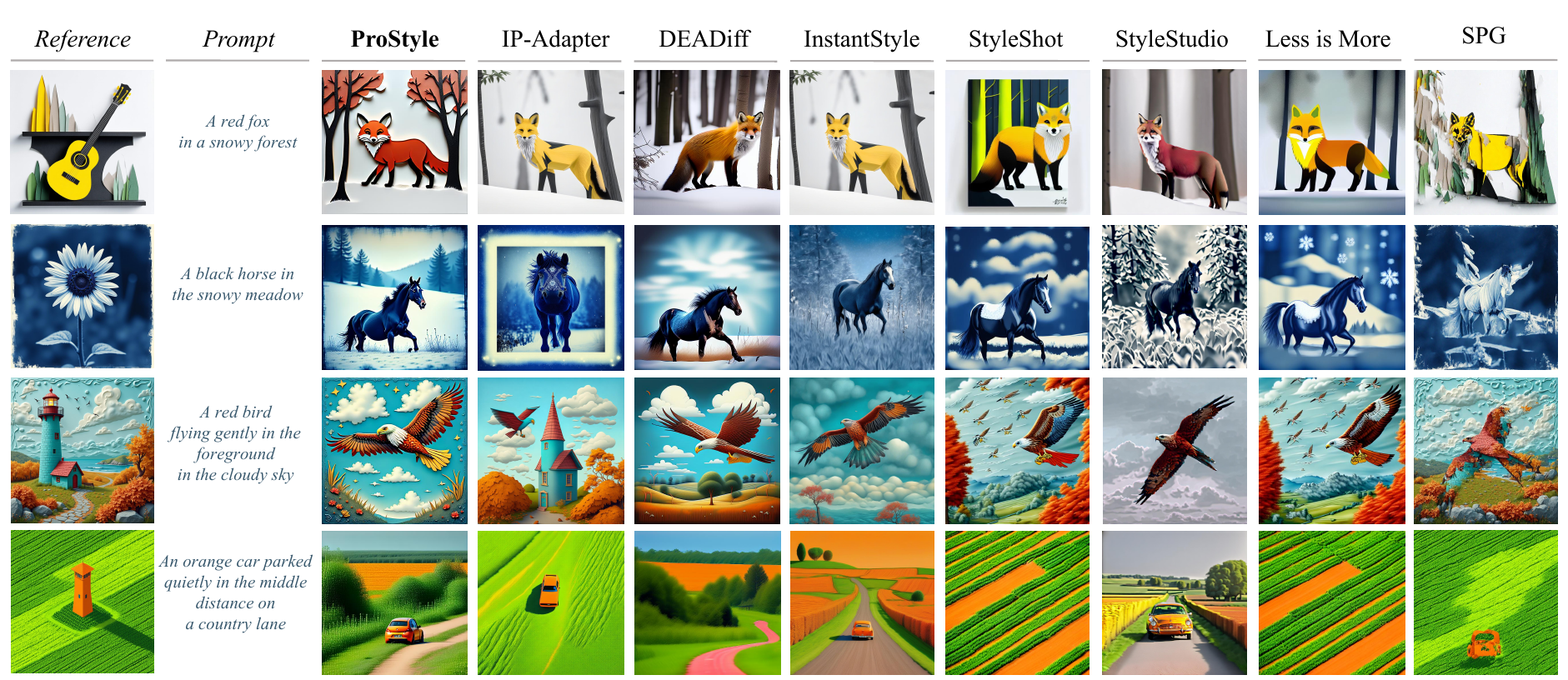}
    \caption{
    Qualitative comparison with seven baselines on ReConStyle-Full. Each row uses the same reference image and target prompt across methods, comparing reference-style preservation with adherence to the requested subjects, colors, and scenes.
    }
    \label{fig:main_exp}
\end{figure*}

Progressive optimization first learns a style prior, then adapts the native pathway, and finally refines both pathways jointly. In Stage 1, we train the Resampler and style-side key and value projections to learn the style prior. In Stage 2, we introduce the native pathway with zero-initialized output projections and train it with the style pathway frozen to learn residual corrections. In Stage 3, we jointly refine both pathways to coordinate their conditioning. The U-Net, VAE, text encoder, and CLIP image encoder remain frozen throughout, and detailed optimizer settings are given in Appendix~\ref{sec:supp_prostyle_training}. All three stages share the same cross-content tuples and minimize the same noise-prediction objective~\cite{ho2020denoising}
\begin{equation}
\mathcal{L}_{\mathrm{CC}}(\psi,\eta)
=\mathbb{E}_{\mathcal{D}_{\mathrm{CC}},t,\epsilon}
\bigl[\|\epsilon-f_{\psi,\eta}(z_t,t,c_t,I_r)\|_2^2\bigr],
\label{eq:cross_content_loss}
\end{equation}
where $\psi$ and $\eta$ denote the style-pathway and native-pathway parameters, and $z_t=\sqrt{\bar\alpha_t}\,z_0+\sqrt{1-\bar\alpha_t}\,\epsilon$ is the noisy target latent at timestep $t$, with $z_0=\mathcal{E}(I_t)$ encoded by the VAE, $\epsilon\sim\mathcal{N}(0,\mathbf{I})$, and $\bar\alpha_t$ given by the noise schedule. No additional style-consistency or reconstruction loss is used.

\subsection{Analysis of the Training Design}
\label{sec:prostyle_theory_main}

We analyze the two pathways through the mean squared denoising risk $\mathcal{R}(f)=\mathbb{E}\|\epsilon-f\|_2^2$ of a noise predictor $f$. For the style pathway, let $N$ denote the reference-specific subjects and background, $X$ the remaining inputs including the shared style, and $\bar f(X)=\mathbb{E}[f(X,N)\mid X]$ the prediction averaged over reference content.

\begin{proposition}[Cross-content risk penalty]
\label{thm:prostyle_nuisance}
For square-integrable $\epsilon$ and $f$, if $\mathbb{E}[\epsilon\mid X,N]=\mathbb{E}[\epsilon\mid X]$, then
$\mathcal{R}(f)-\mathcal{R}(\bar f)=\mathbb{E}\|f-\bar f\|_2^2$.
\end{proposition}

Under conditional mean independence, Proposition~\ref{thm:prostyle_nuisance} quantifies the denoising-risk penalty associated with prediction fluctuations over reference content at fixed $X$. Cross-content pairing motivates this assumption by varying subjects and backgrounds within each style.

For the dual-path analysis, set $U=(z_t,t,c_t)$. With fixed reference mappings, let $f_0(U,S)$ be the trained Stage-1 predictor with style tokens $S$ and $f_\eta(U,S,G)$ the dual-path predictor with native features $G$. Define $m_S=\mathbb{E}[\epsilon\mid U,S]$ and $m_{SG}=\mathbb{E}[\epsilon\mid U,S,G]$.

\begin{proposition}[Dual-path risk decomposition]
\label{thm:prostyle_dual_path_gain}
For square-integrable $\epsilon$, $f_0$, and $f_\eta$,
\begin{equation*}
\begin{aligned}
\mathcal{R}(f_0)-\mathcal{R}(f_\eta)
={}&\mathbb{E}\|m_{SG}-m_S\|_2^2+\mathbb{E}\|m_S-f_0\|_2^2\\
&-\mathbb{E}\|f_\eta-m_{SG}\|_2^2.
\end{aligned}
\end{equation*}
\end{proposition}

With the style pathway fixed, zero-initialized native outputs preserve this predictor and include it in the dual-path function class, whose best attainable denoising risk is therefore no higher. Appendix~\ref{sec:supp_theory} provides the proofs and setup.

\section{Experiments}
\label{sec:experimental}
\begin{table*}[t]
	\centering
	\caption{
		Comparison with existing methods on ReConStyle-Full and in the user study. User-study results are selection rates. Avg. Rank averages ranks across the seven criteria. Bold and underline indicate the best and second-best values. Methods are grouped by backbone (Appendix~\ref{sec:supp_baselines}). Among methods sharing the SD1.5 backbone, ProStyle is best on all four automatic metrics.
	}
	\label{tab:main_comparison}
	\small
	\setlength{\tabcolsep}{5pt}
	\begin{tabular}{llc|cccc|ccc}
		\toprule
		& & & \multicolumn{4}{c|}{ReConStyle-Full} & \multicolumn{3}{c}{User Study (\%)} \\
		\cmidrule(lr){4-7}\cmidrule(lr){8-10}
		Method & Backbone & Avg. Rank $\downarrow$ & Style-CSD $\uparrow$ & CLIP-T $\uparrow$ & TCF $\uparrow$ & ROO $\downarrow$ & Style $\uparrow$ & Text $\uparrow$ & Suitability $\uparrow$ \\
		\midrule
		IP-Adapter   & SD3.5-L & 5.14 & \textbf{0.4101} & 0.2158 & 0.6127 & 0.1782 & \textbf{50.51} & 17.31 & 10.77 \\
		\midrule
		InstantStyle & SDXL & 3.57 & \underline{0.4092} & 0.2553 & 0.7043 & 0.0463 & 36.92 & 36.92 & \underline{14.87} \\
		StyleStudio  & SDXL & \underline{3.43} & 0.3218 & \textbf{0.2678} & \textbf{0.7903} & \textbf{0.0237} & 17.56 & \underline{53.72} & 14.36 \\
		SPG          & SDXL & 7.43 & 0.3549 & 0.1941 & 0.4196 & 0.0895 & 35.00 & 9.23 & 3.72 \\
		\midrule
		DEADiff      & SD1.5 & 4.29 & 0.3750 & 0.2565 & 0.6596 & 0.0290 & 25.90 & 49.87 & 12.18 \\
		StyleShot    & SD1.5 & 5.57 & 0.3841 & 0.2341 & 0.6495 & 0.0492 & 40.38 & 24.74 & 8.21 \\
		Less is More & SD1.5 & 4.71 & 0.3645 & 0.2399 & 0.6607 & 0.0414 & 41.41 & 29.23 & 8.85 \\
		ProStyle     & SD1.5 & \textbf{1.86} & 0.4076 & \underline{0.2668} & \underline{0.7729} & \underline{0.0246} & \underline{42.31} & \textbf{84.36} & \textbf{50.90} \\
		\bottomrule
	\end{tabular}
\end{table*}
\begin{figure*}[t]
	\centering
	\includegraphics[width=\textwidth]{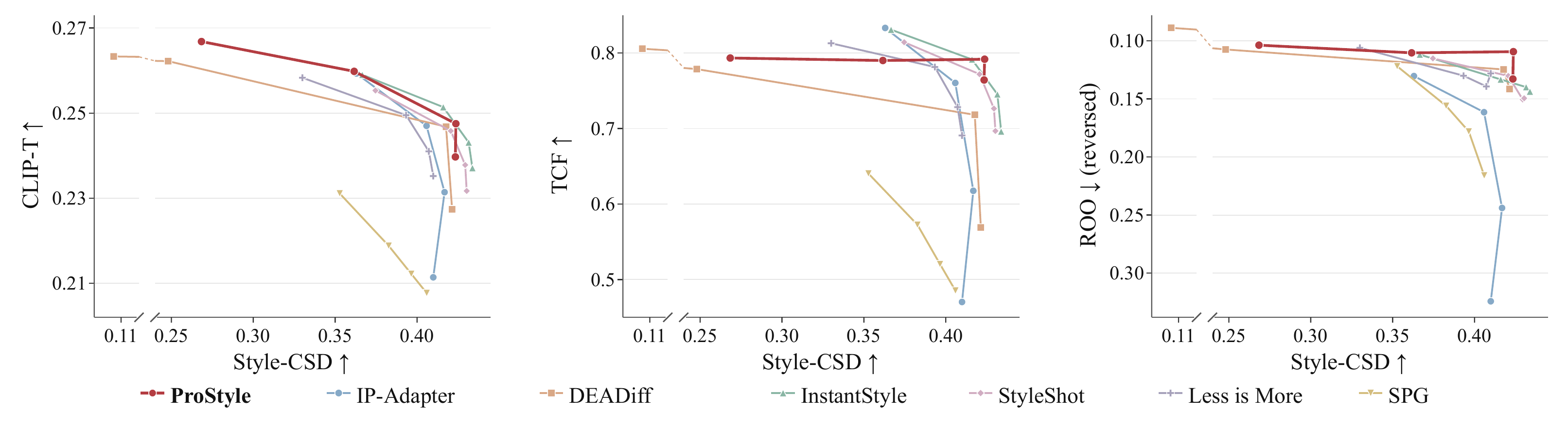}
	\caption{Reference-conditioning strength sweeps on ReConStyle-Rapid. Each curve connects results at $\alpha\in\{0.5,0.75,1.0,1.25\}$, starting from $\alpha=0.5$ at its left end, and shows Style-CSD against CLIP-T (left), TCF (middle), and ROO (right). For ProStyle, $\alpha$ scales all layer weights proportionally. The ROO axis is reversed. The horizontal axis is broken between 0.11 and 0.25 to include DEADiff, and the dashed segment crosses the break. Full results are provided in Appendix~\ref{sec:supp_strength_sweep}.}
	\label{fig:strength_curves}
\end{figure*}
\subsection{Experimental Setup}
\label{sec:experimental_setup}

\paragraph{Implementation Details.}
We use Stable Diffusion v1.5~\cite{rombach2022high} and CLIP ViT-L/14~\cite{radford2021learning}. The three-stage schedule in Section~\ref{sec:progressive_reference_learning} uses AdamW~\cite{loshchilov2017decoupled}, batch size 8, and BF16 on one NVIDIA RTX 5880 Ada GPU. Inference uses 50-step DDIM sampling~\cite{song2020denoising}, classifier-free guidance~\cite{ho2022classifier} scale 8, and image-conditioning scale 1. Automatic metrics are averaged over five inference seed configurations on ReConStyle-Full and three on ReConStyle-Rapid. Appendix~\ref{sec:supp_implementation} gives further details.

\paragraph{Baselines.}
We compare seven methods using official implementations and released weights, including IP-Adapter~\cite{ye2023ip}, DEADiff~\cite{qi2024deadiff}, InstantStyle~\cite{wang2024instantstyle}, StyleShot~\cite{gao2025styleshot}, StyleStudio~\cite{lei2025stylestudio}, Less is More~\cite{zhu2025less}, and SPG~\cite{liang2025spg}. All methods receive the same ReConStyle-Full references and prompts, with inference configurations detailed in Appendix~\ref{sec:supp_baselines}. DEADiff, StyleShot, and Less is More share our SD1.5 backbone and thus provide controlled comparisons, whereas InstantStyle, StyleStudio, and SPG use SDXL~\cite{podell2024sdxl} and IP-Adapter uses SD3.5-Large~\cite{stabilityai2024sd35} (Table~\ref{tab:supp_baseline_configuration}). We report Style-CSD, CLIP-T, TCF, and ROO from Section~\ref{sec:evaluation_metrics}.
\begin{table}[t]
	\centering
	\caption{
		Effect of training-pair construction on ReConStyle-Full.
	}
	\label{tab:cross_content_pairing}
	\small
	\setlength{\tabcolsep}{3.5pt}
	\begin{tabular}{lcccc}
		\toprule
		Training Pairing & Style-CSD$\uparrow$ & CLIP-T$\uparrow$ & TCF$\uparrow$ & ROO$\downarrow$ \\
		\midrule
		Self-Paired & \textbf{0.4549} & 0.2247 & 0.6416 & 0.0938 \\
		Cross-Content & 0.4076 & \textbf{0.2668} & \textbf{0.7729} & \textbf{0.0246} \\
		\bottomrule
	\end{tabular}
\end{table}
\subsection{Comparison with Existing Methods}
\label{sec:comparison_with_existing_methods}

\paragraph{Quantitative Comparison.}
Table~\ref{tab:main_comparison} groups methods by backbone. Under the shared SD1.5 backbone, ProStyle outperforms DEADiff, StyleShot, and Less is More on all four automatic metrics. Methods built on stronger SDXL or SD3.5-Large backbones attain higher scores only on one side of the style--content trade-off. IP-Adapter and InstantStyle exceed ProStyle in Style-CSD by at most 0.0025 but show 7.2$\times$ and 1.9$\times$ its ROO, while StyleStudio leads the three content metrics with a Style-CSD 0.0858 lower. ProStyle is the only method that ranks in the top three on all four automatic metrics, and it achieves the best average rank of 1.86 across the seven evaluation criteria. These comparisons highlight ProStyle's balance between reference-style preservation and target-content control. On the 10 unseen styles, it leads on CLIP-T and ROO and ranks second on TCF, showing that its content-control performance extends to styles excluded from training (Appendix~\ref{sec:supp_style_generalization}). On the external Style-Rank benchmark, whose references are not drawn from ReConStyle, ProStyle achieves the highest CLIP-Text and ImageReward scores among all compared methods, showing that its capabilities generalize beyond our benchmark. Style-Rank's CLIP-Image and DINOv2 scores measure similarity to the reference image and therefore rise with reference-content leakage as well as style similarity. IP-Adapter and SPG lead on these two metrics and also have the two highest ROO values on the same outputs, and IP-Adapter has the lowest CLIP-Text score (Appendix~\ref{sec:supp_style_rank}).

\paragraph{Conditioning Strength.}
Figure~\ref{fig:strength_curves} shows that stronger reference conditioning generally increases style similarity for the baselines while reducing prompt adherence and target-content fidelity and increasing reference-object overlap. By contrast, increasing $\alpha$ from 0.5 to 1.0 raises ProStyle's Style-CSD from 0.2684 to 0.4237, while TCF remains between 0.7900 and 0.7933. This demonstrates that stronger reference conditioning enhances stylistic expression while preserving target-subject fidelity over this range. ProStyle thus reaches a Style-CSD comparable to the strongest baselines while achieving higher CLIP-T and TCF and lower ROO.

\begin{table*}[t]
	\centering
	\caption{
		Component ablation on ReConStyle-Full. The first row uses zero auxiliary scale. Bold indicates the best result in each column.
	}
	\label{tab:ablation}
	\normalsize
	\setlength{\tabcolsep}{6pt}
	\begin{tabular*}{\textwidth}{@{\extracolsep{\fill}}lcccc@{}}
		\toprule
		Variant & Style-CSD$\uparrow$ & CLIP-T$\uparrow$
		& TCF$\uparrow$ & ROO$\downarrow$ \\
		\midrule
		Content-suppressed style pathway
		& 0.3529 & 0.2657 & 0.7733 & \textbf{0.0215} \\
		+ Weak auxiliary-layer conditioning
		& 0.3811 & 0.2645 & 0.7746 & 0.0250 \\
		+ Native reference pathway
		& 0.3996 & \textbf{0.2669} & \textbf{0.7756} & 0.0251 \\
		+ Progressive optimization
		& \textbf{0.4076} & 0.2668 & 0.7729 & 0.0246 \\
		\bottomrule
	\end{tabular*}
\end{table*}
\paragraph{Qualitative Comparison.}
Figure~\ref{fig:main_exp} highlights how reference appearance can conflict with the requested content. In the guitar-to-fox case, most baselines reproduce the reference's yellow on the fox, and DEADiff renders a more naturalistic fox with weaker papercraft texture, whereas ProStyle retains layered paper edges while depicting the requested red fox. In the lighthouse-to-bird case, IP-Adapter carries a tower-like building into the output, while StyleStudio avoids this object but moves toward photographic feathers and clouds. ProStyle retains the reference's raised texture and decorative cloud shapes without reproducing the lighthouse. In the country-lane case, DEADiff, StyleShot, and Less is More omit the requested car, whereas ProStyle retains it. Across these examples, ProStyle mainly improves subject presence, texture transfer, and the suppression of reference objects.
\subsection{Ablation Study}
\label{sec:ablation_study}

\paragraph{Effect of Cross-Content Pairing.}
We replace each training reference with its target image while keeping the architecture, target samples, and training budget fixed. Table~\ref{tab:cross_content_pairing} shows that self-pairing yields higher Style-CSD but raises ROO from 0.0246 to 0.0938 and lowers TCF from 0.7729 to 0.6416. When the reference already contains the target content, the model can fit the target by reusing it, and cross-content pairing limits this route, consistent with Proposition~\ref{thm:prostyle_nuisance}. With the same training data and budget, retrained DEADiff reaches Style-CSD and TCF of 0.3492 and 0.7141, compared with 0.4076 and 0.7729 for ProStyle (Appendix~\ref{sec:supp_self_pair_ablation}).

\begin{figure}[t]
\centering
\includegraphics[width=\linewidth]{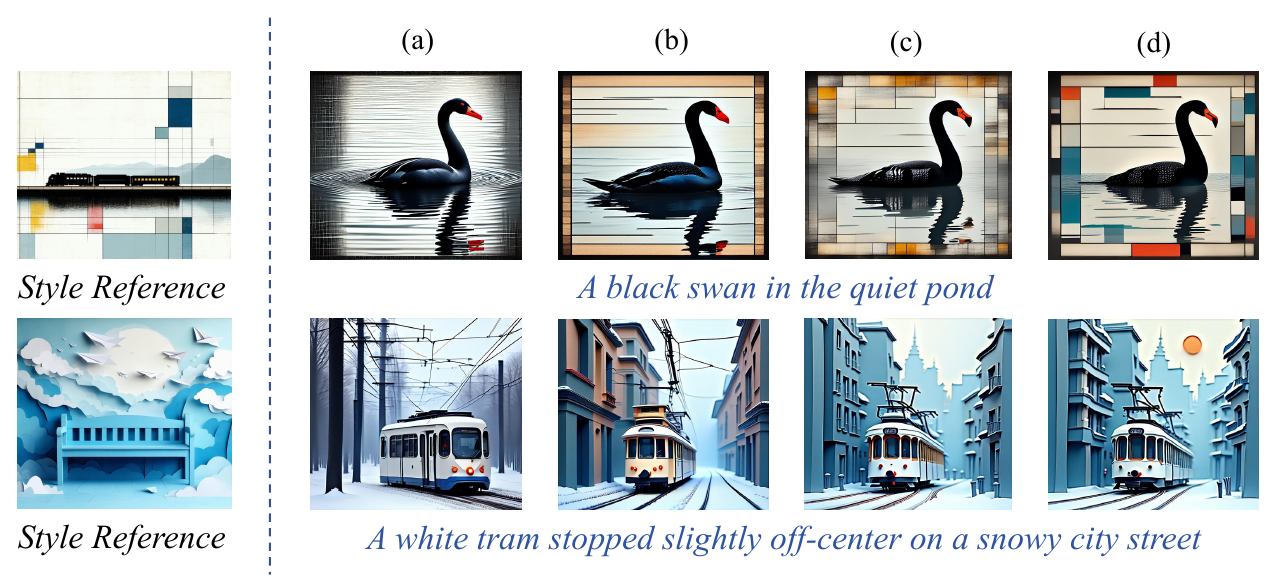}
\caption{
Qualitative ablation corresponding to the four rows of Table~\ref{tab:ablation}, from (a) to (d). Each component adds reference-style characteristics while preserving the requested subjects.
}
\label{fig:ablation}
\end{figure}

\paragraph{Component Ablation.}
Table~\ref{tab:ablation} follows the order of the method. Weak auxiliary-layer conditioning raises Style-CSD from 0.3529 to 0.3811 with stable TCF. The native pathway, which adds full reference features, raises it to 0.3996, and progressive optimization reaches 0.4076 under the same architecture. Across these steps, TCF stays between 0.7729 and 0.7756 and ROO stays at or below 0.0251, so the style gains come without loss of content control. Figure~\ref{fig:ablation} shows corresponding gains in reference-style characteristics while preserving the requested subjects.

\paragraph{Targeted Controls.}
\label{sec:targeted_controls}
Appendix~\ref{sec:supp_component_ablation} reports additional controls under matched training budgets. Strong auxiliary conditioning only slightly raises Style-CSD over the weak setting, but lowers TCF and raises ROO. Compared with native-only conditioning, jointly trained dual pathways raise Style-CSD from 0.3863 to 0.3996 and TCF from 0.7551 to 0.7756, showing that the style prior complements full reference features. Low-learning-rate continuation of the style pathway reaches Style-CSD 0.3731, indicating that additional style-pathway training alone does not reproduce the dual-pathway gain. Extended Stage 2 replaces joint refinement with continued native-only training, and ProStyle improves over it in Style-CSD from 0.3998 to 0.4076 and TCF from 0.7667 to 0.7729 while reducing ROO by 23\% from 0.0318 to 0.0246. Thus, extending native-only training does not reproduce the benefits of joint refinement.

\paragraph{Backbone Transfer.} In a preliminary transfer without backbone-specific layer selection, ProStyle raises CLIP-T over SD1.5 on SD2.1, SDXL, and SD3.5-Large, and SDXL and SD3.5-Large also improve TCF and ROO on ReConStyle-Rapid (Appendix~\ref{sec:supp_backbone_transfer}).

\subsection{User Study}
\label{sec:user_study}

To assess perceptual transfer, 26 participants each evaluate the same 30 cases with anonymous outputs from ProStyle and seven baselines. They select results for style consistency, text alignment, and combined suitability. Combined suitability considers prompt adherence, reference-style similarity, and visible generation defects (Appendix~\ref{sec:supp_user_study}).

Table~\ref{tab:main_comparison} shows that ProStyle ranks second in style and first in text alignment and combined suitability, with selection rates of 84.36\% and 50.90\% that exceed the next-best methods by 30.64 and 36.03 percentage points, respectively. These results highlight ProStyle's ability to transfer reference appearance while following the requested content.

\section{Conclusion}
\label{sec:conclusion}

We present \textbf{ReConStyle} and \textbf{ProStyle} to balance reference-guided style transfer with control over text-specified content. ReConStyle provides 134K cross-content pairs, two test settings, and complementary measures of style similarity and content control. We will release the dataset and its evaluation code to support further research on reference-guided generation. On these pairs, ProStyle learns a content-suppressed style prior, extends it to auxiliary layers through weak conditioning to strengthen style while preserving content, and integrates full reference features through residual fitting followed by joint refinement. ProStyle obtains the best average rank on ReConStyle-Full and leads the user study in text alignment and combined suitability.
\section{LLM Usage Statement}
\label{sec:llm_usage_statement}

We used large language model (LLM)-based assistants throughout this work. They supported the literature survey and reference search, drafted and debugged code for training, evaluation, and data processing, assisted in designing experiments and analyzing results, and drafted and polished the manuscript. The authors reviewed and verified all LLM-assisted content, including code, references, results, and text, and take full responsibility for this paper.
{
    \small
    \bibliographystyle{ieeenat_fullname}
    \bibliography{main}
}

\clearpage
\setcounter{page}{1}
\maketitlesupplementary
\appendix

\section{Related Work}
\label{sec:related_work}

This section expands on the related work discussed in Section~\ref{sec:introduction}.

\subsection{Text-to-Image Generation}
\label{sec:text_to_image_generation}
Diffusion models enable high-quality text-conditioned synthesis~\cite{rombach2022high,saharia2022photorealistic}. Textual Inversion and StyleDrop extend this control to user-provided visual concepts and styles through per-concept optimization~\cite{gal2022image,sohn2023styledrop}. Reference-guided stylization instead conditions generation on a reference image at inference time, supplying appearance information that is difficult to describe in text alone. ProStyle follows this setting and requires no optimization for each new reference.

\subsection{Reference-Guided Stylized Generation}
\label{sec:reference_guided_stylized_generation}
Classical style transfer matches feature statistics or learns separate content and style representations~\cite{gatys2016image,huang2017arbitrary,huang2018multimodal}. Diffusion-based methods introduce the reference through image embeddings or shared and swapped attention~\cite{pan2023arbitrary,ye2023ip,hertz2024style,jeong2024visual}. Subsequent work learns style representations through dedicated encoders and disentanglement objectives~\cite{wang2023stylediffusion,wang2025styleadapter,gao2025styleshot,li2024styletokenizer}, or separates content and style through low-rank adaptation~\cite{frenkel2024implicit,liu2025unziplora}. Selective conditioning and guidance provide further control over transferred style elements~\cite{lei2025stylestudio,wang2024instantstyle}. StyleMaster extends reference-guided stylization to video~\cite{ye2025stylemaster}. USO jointly models subject- and style-driven generation with an auxiliary style reward~\cite{wu2026unified}.

DEADiff learns style representations from pairs of images generated with the same prompt, which therefore share both style and subject, and injects them into selected attention layers~\cite{qi2024deadiff}, whereas ProStyle pairs images that share only style and differ in subject and background, adds weak conditioning at auxiliary layers, and progressively learns complementary residuals from full reference features. CSGO conditions on separate content and style images and also supports text-driven stylized generation~\cite{xing2024csgo}. ProStyle uses a single style reference and represents it at two levels of compression, through 16 style tokens and the full patch features.

\subsection{Leakage Control}
\label{sec:leakage_control}
Several methods explicitly target the transfer of unwanted reference content. Less is More masks content-associated feature elements, and StyleKeeper and Only-Style regulate leakage during generation~\cite{zhu2025less,jeong2025stylekeeper,aravanis2025only}. CleanStyle applies denoising-dependent feature purification, and SafeStyle combines calibrated subspaces with budgeted residual injection~\cite{feng2026cleanstyle,yang2026safestyle}. These methods also recognize that suppressing content can remove transferable style. ProStyle addresses this issue during training. It first learns a content-suppressed style pathway under cross-content supervision and then fits residual corrections from the full reference features while this pathway is frozen.

\subsection{Staged Learning}
\label{sec:staged_learning}
UniCSG uses stage-specific objectives for semantic disentanglement and frequency-aware reconstruction, together with pixel-space reward learning~\cite{yang2026unicsg}. ProStyle instead retains one cross-content noise-prediction objective throughout training, first learning the style pathway, then fitting native residuals with the style pathway frozen, and finally refining both pathways jointly.

\subsection{Evaluation of Stylized Generation}
\label{sec:related_evaluation}
Stylized generation is commonly evaluated by computing similarity to the reference image with CLIP~\cite{radford2021learning}, DINOv2~\cite{DBLP:journals/tmlr/OquabDMVSKFHMEA24}, or CSD~\cite{somepalli2024investigating} features, and image--text alignment with CLIP-T~\cite{radford2021learning}. These metrics provide limited insight into fine-grained subject and attribute correctness and the leakage of reference objects into generated images. TIFA evaluates text-to-image faithfulness through visual question answering~\cite{hu2023tifa}, while GenEval assesses object-level alignment, including object presence, count, color, and position~\cite{ghosh2023geneval}. Davidsonian Scene Graph improves the reliability of fine-grained evaluation by accounting for dependencies among evaluation questions~\cite{cho2024davidsonian}. VQAScore measures image--text alignment using the probability that a visual question-answering model answers affirmatively to a question about the image and text~\cite{lin2024evaluating}. ReConStyle complements these alignment-oriented evaluations with VLM-based object-level measures of target-content fidelity and non-target reference-object overlap~\cite{bai2025qwen3}.

\begin{figure*}[t]
    \centering
    \includegraphics[width=\textwidth]{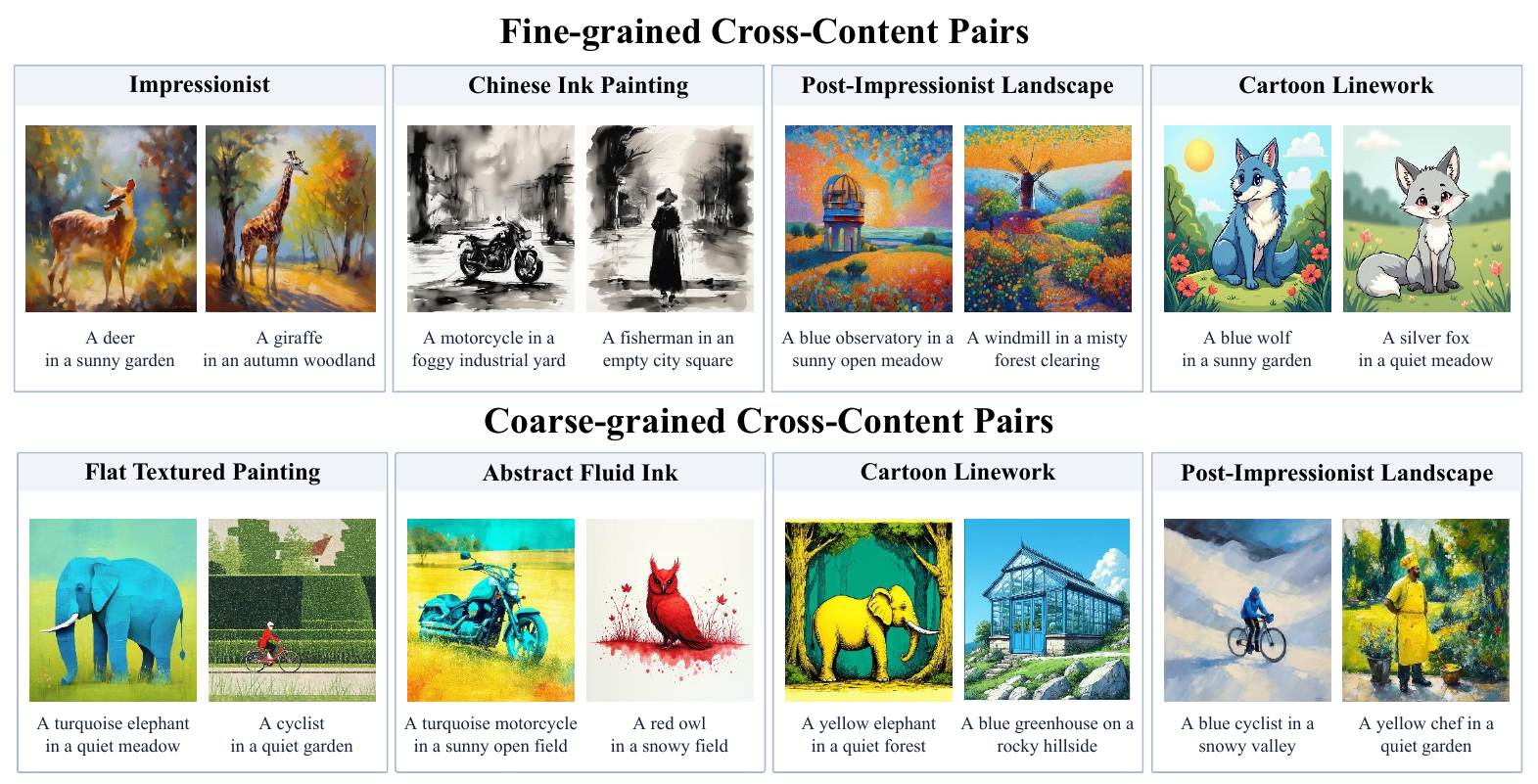}
    \caption{
    Representative fine-grained and coarse-grained cross-content training pairs from ReConStyle.
    Fine-grained pairs exhibit higher stylistic similarity (CSD $>0.40$), while coarse-grained pairs capture broader stylistic variations (CSD $<0.35$).
    Each image is shown with its content prompt.
    }
    \label{fig:supp_training_pairs}
\end{figure*}

\begin{table*}[t]
\centering
\caption{
Style categories and their assignments in ReConStyle. Category names follow the listed ID order, and unseen styles are excluded from training.
}
\label{tab:supp_style_categories}
\small
\setlength{\tabcolsep}{6pt}
\begin{tabular}{p{0.15\textwidth}lp{0.66\textwidth}}
\toprule
Assignment & Style IDs & Style Categories \\
\midrule
Training and Full & S01--S10 &
Impressionist, Expressive Watercolor, Chinese Ink Painting,
Ukiyo-e, Papercut Collage, Oil Painting,
Dense Blue Textured Painting, Pencil Sketch,
Thick Impasto, Pointillism \\

& S11--S20 &
Mosaic Texture, Cyanotype, Piled-Color Brushwork,
Flat Textured Painting, Light Watercolor,
Bauhaus Geometry, Traditional Pattern,
Textured Oil Abstraction, Abstract Fluid Ink,
Vector Pattern \\

& S21--S30 &
Textile Ornament, Geometric Abstract,
Abstract Expression, Marbled Color Flow,
Monochrome Woodcut, Minimalist Scene,
Op Art, Abstract Watercolor,
Watercolor Landscape, Watercolor Scenic \\

& S31--S40 &
Quiet Oil Landscape, Gradient Scenic,
Cubist Abstraction, Soft Painted Landscape,
Mondrian Geometry, Retro Decorative Flat,
Abstract Mosaic, Embossed Relief,
Monochrome Ornamental, Post-Impressionist Landscape \\

\midrule
Training only & S41--S50 &
Dark Engraving, Cartoon Linework,
Japanese Manga, Sci Fi Manga,
Pop Art, Art Nouveau,
Mural Painting, Cyberpunk City,
Fantasy Illustration, Isometric Art \\
\midrule
Full only (unseen) & G01--G10 &
Stained Glass, Gongbi Painting, Chinese Baimiao,
Baroque Painting, Rococo Painting, Steampunk,
Vaporwave, Retro Pixel Game Art,
Stylized 3D Cartoon, Fauvism \\
\bottomrule
\end{tabular}
\end{table*}

\section{ReConStyle Benchmark Details}
\label{sec:supp_benchmark}

ReConStyle provides cross-content training pairs for learning a content-suppressed style prior and evaluates style similarity, text alignment, target-content fidelity, and non-target reference-object overlap.

\subsection{Style Categories and Data Sources}
\label{sec:supp_style_categories}

\paragraph{Style Categories.}
Training covers 50 visual style categories. Each category has eight fine-grained textual descriptions, giving 400 training style profiles. These descriptions vary brushwork, contours, textures, shading, and color treatment within a category. ReConStyle-Full contains 50 categories, including 40 shared with training and 10 unseen styles. Training and Full share S01--S40. S41--S50 appear only in training, and G01--G10 appear only in Full. Table~\ref{tab:supp_style_categories} lists their assignments.

\paragraph{Training Image Sources.}
The training collection contains 21,374 images, including 8,000 generated by FLUX.1-dev~\cite{blackforestlabs2024flux}, 8,000 by Stable Diffusion 3.5 Large~\cite{stabilityai2024sd35}, and 5,374 supplementary images obtained using Qwen-Image-Edit 2511~\cite{wu2025qwenimage}. For the two text-to-image generators, each generation prompt combines a content description with a fine-grained style description. We use Qwen-Image-Edit to supplement style profiles with fewer eligible pairing candidates. The editing instruction specifies the desired style while retaining the source content and composition. Images from all three sources enter a shared pairing pool, so a reference and its target may come from different generators.

\subsection{Cross-Content Training Set Construction}
\label{sec:supp_training_set}

\paragraph{Image and Text Organization.}
Each training image is associated with structured metadata describing its content and style, including a style category, a fine-grained style profile, a content prompt, and identifiers for the subject and background. During training, the target prompt $c_t$ contains only the content description of the target image $I_t$, excluding the style description used during image generation. For Qwen-edited images, the content description is inherited from the source image. Each training example is therefore a tuple $(I_r,I_t,c_t)$, where $I_r$ provides the reference style and $c_t$ specifies the target content.

\paragraph{Cross-Content Pairing.}
We construct training pairs within the same style category while enforcing differences in reference and target content. Specifically, a candidate pair is valid only when the two images have different content identifiers, subject keys, and background keys, as determined by their structured metadata. These constraints vary content within a style category to discourage dependence on reference-specific semantics and support a transferable style prior. An image may participate in multiple training pairs, but each unordered image pair is selected at most once.

\paragraph{CSD-Based Pair Selection.}
To capture different degrees of stylistic similarity within each category, we divide eligible pairs into fine-grained and coarse-grained groups using CSD similarity~\cite{somepalli2024investigating}. Fine-grained pairs have a CSD similarity above 0.40, while coarse-grained pairs have a similarity below 0.35. Candidates with similarities between 0.35 and 0.40 are excluded. The two groups provide closely matched appearances and broader stylistic variation within each category, as illustrated in Figure~\ref{fig:supp_training_pairs}.

\paragraph{Balanced Pair Sampling.}
We apply balanced sampling across style categories to avoid disproportionate contributions from individual styles. Each of the 50 categories contributes 1,876 fine-grained pairs and 804 coarse-grained pairs, maintaining a 70:30 ratio. Candidate selection considers both CSD similarity and image reuse frequency, with each image appearing as a reference in at most nine selected pairs. The resulting training set contains 93,800 fine-grained and 40,200 coarse-grained pairs, totaling 134,000 cross-content training pairs, as summarized in Table~\ref{tab:supp_pair_statistics}.

\paragraph{Quality Control.}
For each style category, we randomly inspect 50 selected pairs for compatible stylistic appearances despite their different contents. Each of the 50 categories achieves an acceptance rate of at least 90\% in this sample-level check. The inspection supports the stylistic compatibility of the sampled pairs.

\begin{table}[t]
\centering
\caption{
Statistics of the cross-content training pairs.
}
\label{tab:supp_pair_statistics}
\small
\setlength{\tabcolsep}{4pt}
\begin{tabular}{lcccc}
\toprule
Pair Type & CSD Criterion & Per Style & Total & Ratio \\
\midrule
Fine-grained & $>0.40$ & 1,876 & 93,800 & 70\% \\
Coarse-grained & $<0.35$ & 804 & 40,200 & 30\% \\
\midrule
Total & -- & 2,680 & 134,000 & 100\% \\
\bottomrule
\end{tabular}
\end{table}

\subsection{Test Set Construction}
\label{sec:supp_test_set}

\paragraph{ReConStyle-Rapid.}
ReConStyle-Rapid supports fast validation during development and ablation. It contains 100 synthetic references from 10 style categories, with 50 images generated by FLUX.1-dev and 50 by Stable Diffusion 3.5 Large. For each reference, we construct three progressive prompts (P1--P3) while keeping the reference image and target subject fixed. P1 specifies the target subject, P2 adds a subject attribute, and P3 further incorporates scene and background constraints. Within each inference seed configuration, the same seed is used across the three prompt levels for each reference. This yields 300 evaluation conditions.

\paragraph{ReConStyle-Full.}
ReConStyle-Full provides a larger evaluation set with more diverse references and structured target descriptions. It covers 50 style categories, including 40 seen categories shared with the training set and 10 unseen styles excluded from training. Each category contains 28 FLUX.1-dev references, 28 Stable Diffusion 3.5 Large references, and 10 images manually selected from Style30K. This yields 3,300 reference images. The synthetic references are constructed using four fine-grained style descriptions per category, each combined with seven content conditions and generated by both models. Each reference is paired with two independently constructed prompts P3 and P4, producing 6,600 evaluation conditions. P3 specifies a target subject, subject attribute, scene attribute, and background. P4 includes an action description and a spatial placement constraint, and Figure~\ref{fig:dataset_benchmark} shows its action component. Unlike the progressive prompts in Rapid, the two prompts are constructed independently and need not share the same target subject or scene. Table~\ref{tab:supp_full_statistics} summarizes the seen and unseen subsets.

\begin{table}[t]
\centering
\caption{Statistics of ReConStyle-Full. References reports reference-image counts. Each reference is paired with two target prompts.}
\label{tab:supp_full_statistics}
\small
\setlength{\tabcolsep}{3pt}
\begin{tabular}{lrrr}
\toprule
Split & Styles & References & Conditions \\
\midrule
Seen & 40 & 2,640 & 5,280 \\
Unseen & 10 & 660 & 1,320 \\
\midrule
Full & 50 & 3,300 & 6,600 \\
\bottomrule
\end{tabular}
\end{table}

\begin{table}[t]
	\centering
	\caption{
		Representative target prompts from ReConStyle-Rapid and ReConStyle-Full.
		Rapid progressively adds constraints to the same target subject, while Full uses independently constructed prompts.
	}
	\label{tab:supp_prompt_examples}
	\small
	\setlength{\tabcolsep}{4pt}
	\begin{tabular}{llp{0.62\linewidth}}
		\toprule
		Setting & Level & Target Prompt \\
		\midrule
		Rapid & P1 & An owl. \\
		Rapid & P2 & A white owl. \\
		Rapid & P3 & A white owl in a moonlit forest. \\
		\midrule
		Full & P3 & A red-coated hiker on a foggy trail. \\
		Full & P4 & An orange guitar leaning gently near the center of the frame under the soft lamplight. \\
		\bottomrule
	\end{tabular}
\end{table}

\paragraph{Content Separation.}
We apply content-level constraints to reduce overlap between training and evaluation conditions. The evaluation prompts are checked against the training content pool using complete content descriptions and normalized subject--background combinations. No exact content-prompt overlap or normalized subject--background combination overlap is found in the final training collection, so each evaluation condition pairs its subject and background in a combination unseen in training. Because synthetic references are generated from known content descriptions, their subjects are additionally required to differ from the target subjects. Representative prompts from both evaluation settings are provided in Table~\ref{tab:supp_prompt_examples}.

\section{Evaluation Protocol}
\label{sec:supp_evaluation}

\subsection{Style Consistency and Text Alignment}
\label{sec:supp_style_text_metrics}

\paragraph{Style Consistency.}
Directly computing CSD, CLIP, or DINOv2 similarity between generated and reference images may be influenced by their semantic content, particularly the subjects depicted in the reference. To mitigate this influence, we evaluate style consistency using images without salient subjects that represent the corresponding reference style. We use the frozen CSD ViT-L/14 encoder~\cite{somepalli2024investigating} to extract 768-dimensional style embeddings, with images resized and center-cropped to $224\times224$ using CLIP normalization.

We construct eight style profiles per category for ReConStyle-Rapid and 15 for ReConStyle-Full. Each profile contains four subject-free images generated using FLUX.1-dev and Stable Diffusion 3.5 Large, with two random seeds per model. 

Style-CSD is computed as the cosine similarity between the generated image embedding and the style representation derived from these exemplars.

Although this approach reduces the influence of reference-specific subjects, the complexity and diversity of visual styles mean that the resulting scores may not fully reflect human perception of style consistency. We therefore complement Style-CSD with human evaluation, as described in Appendix~\ref{sec:supp_user_study}.

\paragraph{Text Alignment.}
We compute CLIP-T using the frozen CLIP ViT-L/14 model~\cite{radford2021learning}. Generated images are processed at $224\times224$ resolution, and the complete target prompts are encoded using the CLIP text encoder. CLIP-T is calculated as the cosine similarity between the normalized image and text embeddings. Both Style-CSD and CLIP-T use raw cosine similarities without additional scaling.

\subsection{Target-Content Fidelity}
\label{sec:supp_tcf}

\paragraph{Target Content Extraction.}
Qwen3-VL extracts the visible objects and their appearance properties from each generated image alone, excluding broad scene regions and ambiguous visual elements. Each object record contains a noun category, visible colors, patterns or materials, and an instance count. TCF then compares these records with the target subject and its requested appearance attribute, which the evaluation manifest specifies.

\paragraph{Category and Attribute Matching.}
Object categories are normalized through lowercase conversion, simple plural handling, and a fixed alias dictionary, such as mapping \textbf{watchtower} to \textbf{tower}. Matching also accepts detected subcategories of the target, such as \textbf{armchair} for \textbf{chair}, and a detected person for role-based targets such as hikers, chefs, and cyclists. Appearance attributes are checked separately using predefined rules for equivalent and partially matching descriptions of colors, patterns, and materials. Based on these rules, subject presence receives a binary score of 0 or 1, and attribute matching receives a score of 0, 0.5, or 1.

\paragraph{Special Cases.}
A missing target subject, including a valid empty extraction, receives TCF $=0$. A detected target subject with missing or unmatched attributes receives TCF $=0.5$. When the prompt requests no attribute, as in the P1 prompts of ReConStyle-Rapid, $M_{\mathrm{attr}}=1$. Multiple instances of the same category are evaluated using their combined appearance properties.

\subsection{Reference-Object Overlap}
\label{sec:supp_roo}

\paragraph{Reference Object Extraction.}
We extract object categories from reference and generated images using the same VLM-based procedure. ROO measures the proportion of the remaining reference categories that also appear in the generated image, following the definition in Section~\ref{sec:evaluation_metrics}. Each object category is counted once regardless of its instance count. Recognizable objects in the background, such as towers or boats, are included, whereas broad environmental regions, such as sky, water surfaces, fields, and roads, are excluded from both object sets. Because prompt-specified scenes mostly consist of such regions, they rarely contribute to ROO.

\paragraph{Semantic Matching and Validity.}
Category matching follows the same canonicalization and alias rules as TCF. Before computing ROO, target-equivalent categories, including hyponyms and hypernyms such as \textbf{bird} for a crane target, are removed from both object sets, and synonymous categories such as \textbf{stone} and \textbf{rock} are merged. Conditions with no remaining reference objects are excluded from aggregation. A valid empty generated object set receives ROO $=0$.

\subsection{Evaluation Implementation}
\label{sec:supp_evaluation_implementation}

\paragraph{VLM-Based Evaluation.}
We use Qwen3-VL-8B-Instruct~\cite{bai2025qwen3} with BF16 precision and greedy decoding. The model processes each image independently and returns structured object records containing category names, visible appearance attributes, and instance counts. The extraction instructions focus on clearly identifiable physical entities and exclude scene descriptions, actions, spatial relations, and stylistic attributes. Outputs are parsed and normalized before scoring. The same extracted records are reused for TCF and ROO.

\paragraph{Metric Aggregation.}
Metrics are computed for individual evaluation conditions and averaged for each method. Style-CSD and CLIP-T use successfully encoded images. Missing images and extraction failures are excluded from the average. On ReConStyle-Full, all methods have valid Style-CSD, CLIP-T, and TCF scores for all 6,600 conditions. All methods compute ROO on the same set of about 6,300 conditions, excluding only references without non-target objects.

\subsection{Reliability of VLM Extraction}
\label{sec:supp_vlm_reliability}

TCF and ROO are computed in two steps. Qwen3-VL first extracts object categories, attributes, and instance counts from each generated and reference image alone. Deterministic rules then match the extractions against the target subject and the reference objects to compute the scores. Given the extracted records, the second step is deterministic and identical across all methods.

To audit the extraction step, we sample 200 conditions from the ProStyle evaluation on ReConStyle-Full, including 150 drawn uniformly at random and 50 drawn from conditions flagged as reference-content leakage. We further sample 50 conditions from each of three baselines, namely IP-Adapter, StyleStudio, and SPG, including 40 drawn uniformly at random and 10 drawn from conditions flagged as leakage.

Three human annotators independently inspect both the reference and generated images for all 350 conditions, labeling target presence, requested attributes, instance counts, and non-target reference objects. We apply the same deterministic scoring rules to these annotations and compare the resulting TCF and ROO scores with the VLM-based scores. On the 270 randomly sampled audit conditions across four methods, Table~\ref{tab:supp_vlm_audit} shows that the mean agreement across the three annotators is 79.1\% for TCF and 87.3\% for ROO. The 80 additional leakage-targeted conditions are used for supplementary inspection and are not included in these agreement rates.

The inspection identifies two recurring types of deviations. First, without access to the prompt, the VLM sometimes reports a coarser category than the target, such as \textbf{bird} for a swan, which can lower TCF when the fixed alias dictionary does not cover the case. Second, recognizable background or small objects shared by the reference and generated images can raise ROO.

Across eight methods, TCF and human text-alignment rankings agree closely, with Spearman $\rho=0.90$. This method-level comparison complements the condition-level audit above.

\begin{table}[t]
\centering
\caption{Agreement between annotation-derived and VLM-derived TCF and ROO scores on 270 randomly sampled conditions across four methods, averaged across three annotators. Agreement requires identical scores.}
\label{tab:supp_vlm_audit}
\small
\begin{tabular}{ccc}
\toprule
Audit conditions & TCF agreement & ROO agreement \\
\midrule
270 & 79.1\% & 87.3\% \\
\bottomrule
\end{tabular}
\end{table}

\section{Analysis Details and Proofs}
\label{sec:supp_theory}

This section provides the complete setup and proofs for Propositions~\ref{thm:prostyle_nuisance} and~\ref{thm:prostyle_dual_path_gain} in Section~\ref{sec:prostyle_theory_main}.

\subsection{Setup}
\label{sec:supp_theory_setup}

Let $z_t$ denote the noisy target latent, $t$ the diffusion timestep, $c_t$ the target text prompt, and $\epsilon$ the injected Gaussian noise. Set $U=(z_t,t,c_t)$, let $S^\star$ denote a conceptual shared style and $N$ the reference subjects and background, and write $X=(U,S^\star)$. For fixed model parameters, let $f$ be a noise-prediction function in the analytical variables $(X,N)$, with output $f(X,N)$ of the same dimension as $\epsilon$. Expectations $\mathbb E$ average over cross-content training examples, timesteps, and injected noise, and $\|\cdot\|_2$ denotes the Euclidean norm. The mean squared risk functional $\mathcal R$ evaluates each predictor at its indicated inputs. For $f$, define its risk and content-averaged predictor by
\begin{equation}
\begin{aligned}
\mathcal R(f)&=\mathbb E\|\epsilon-f(X,N)\|_2^2,\\
\bar f(X)&=\mathbb E[f(X,N)\mid X].
\end{aligned}
\label{eq:prostyle_risk_and_average}
\end{equation}
The comparator $\bar f(X)$ averages predictions over reference content at fixed $X$. Predictor symbols denote functions in $\mathcal R(\cdot)$ and evaluated outputs within expectations and vector expressions, with input arguments omitted when clear.

\paragraph{Proposition~\ref{thm:prostyle_nuisance} (Cross-content risk penalty, restated).}
\textit{Suppose $\epsilon,f(X,N)$ are square-integrable and $\mathbb E[\epsilon\mid X,N]=\mathbb E[\epsilon\mid X]$. Then}
\begin{equation*}
\mathcal R(f)-\mathcal R(\bar f)
=\mathbb E\|f-\bar f\|_2^2.
\end{equation*}

Proposition~\ref{thm:prostyle_nuisance} quantifies the risk penalty from prediction fluctuations over reference content. The penalty is strictly positive when the mean squared fluctuation is nonzero.

Compact tokens can also omit informative detail, motivating native compensation. Let $F$ denote the frozen CLIP reference patch features, $R_\phi$ the Resampler with parameters $\phi$, and $\operatorname{LN}$ layer normalization. With the reference mappings fixed, write $S=R_\phi(F)$ for the learned tokens and $G=\operatorname{LN}(F)$ for normalized patches. Fix the trained Stage-1 style-pathway predictor $f_0(U,S)$. Here $S$ is the learned representation, whereas $S^\star$ is the conceptual shared style. Let $f_\eta(U,S,G)$ denote the complete dual-path predictor, where $\eta$ comprises the native-pathway parameters and the remaining model parameters are fixed. Both predictors are measurable with respect to their stated inputs. This fixed-baseline setup describes Stage 2. Define
\begin{equation}
m_S=\mathbb E[\epsilon\mid U,S],\qquad
m_{SG}=\mathbb E[\epsilon\mid U,S,G].
\label{eq:prostyle_complementary_means}
\end{equation}
These conditional means minimize squared prediction error for the specified inputs. Both $S$ and $G$ come from the same reference features, and the decomposition requires no independence between them. The next proposition follows from nested conditioning and decomposes the risk difference between the trained predictors.

\paragraph{Proposition~\ref{thm:prostyle_dual_path_gain} (Dual-path risk decomposition, restated).}
\textit{For square-integrable $\epsilon,f_0,f_\eta$,}
\begin{equation*}
\begin{aligned}
\mathcal R(f_0)-\mathcal R(f_\eta)
={}&\mathbb E\|m_{SG}-m_S\|_2^2\\
&+\mathbb E\|m_S-f_0\|_2^2
-\mathbb E\|f_\eta-m_{SG}\|_2^2.
\end{aligned}
\end{equation*}

The first term measures complementary prediction information supplied by $G$ beyond $S$, the second captures approximation error in the trained style pathway, and the third measures the remaining dual-path error. Native adaptation strictly reduces risk when the third term is smaller than the first two combined. Zero-initialized native outputs preserve $f_0$ exactly. With the style pathway frozen during Stage 2, minimizing the existing noise-prediction objective is equivalent to fitting the predictable residual of this baseline.

\subsection{Proof of Proposition~\ref{thm:prostyle_nuisance}}
\label{sec:prostyle_nuisance_proof}

The proof separates reference-content fluctuations from the content-averaged prediction. Conditional mean independence makes the two contributions orthogonal.

\begin{proof}[Proof of Proposition~\ref{thm:prostyle_nuisance}]
Let $m_X=\mathbb E[\epsilon\mid X]$ be the conditional mean of the target noise and $u=f-\bar f$ the centered prediction fluctuation over reference content. Conditional averaging gives
\begin{equation}
\mathbb E[u\mid X]=0.
\label{eq:prostyle_centered_content_fluctuation}
\end{equation}
The variables $u$ and $\bar f$ are measurable with respect to $(X,N)$, while $m_X-\bar f$ is $X$-measurable. Writing $\langle\cdot,\cdot\rangle$ for the Euclidean inner product, the assumed conditional mean independence and the tower property yield
\begin{equation}
\begin{aligned}
\mathbb E\langle\epsilon-\bar f,u\rangle
&=\mathbb E\langle m_X-\bar f,u\rangle\\
&=\mathbb E\langle m_X-\bar f,\mathbb E[u\mid X]\rangle=0.
\end{aligned}
\label{eq:prostyle_content_orthogonality}
\end{equation}
All inner products are integrable by Cauchy--Schwarz. Since $\epsilon-f=(\epsilon-\bar f)-u$, expansion gives
\begin{equation}
\begin{aligned}
\mathcal R(f)
&=\mathcal R(\bar f)+\mathbb E\|u\|_2^2
-2\mathbb E\langle\epsilon-\bar f,u\rangle\\
&=\mathcal R(\bar f)+\mathbb E\|f-\bar f\|_2^2.
\end{aligned}
\label{eq:prostyle_content_risk_proof}
\end{equation}
Subtracting $\mathcal R(\bar f)$ proves the proposition.
\end{proof}

\subsection{Proof of Proposition~\ref{thm:prostyle_dual_path_gain}}
\label{sec:prostyle_dual_path_proof}

We first separate the irreducible prediction error from approximation error using nested conditioning. This yields the risk difference formula for the actual Stage-1 predictor. We then express native adaptation as residual fitting and verify preservation at zero initialization.

\begin{proof}[Proof of Proposition~\ref{thm:prostyle_dual_path_gain}]
Let $\sigma(\cdot)$ denote the sigma-algebra generated by its arguments. For fixed reference representations, set $\mathcal A=\sigma(U,S)$ and $\mathcal B=\sigma(U,S,G)$, representing the information available to the style-pathway and dual-path predictors, respectively. Then $\mathcal A\subseteq\mathcal B$ and
\begin{equation}
\mathbb E[m_{SG}\mid\mathcal A]=m_S,
\qquad \mathbb E[m_{SG}-m_S\mid\mathcal A]=0.
\label{eq:prostyle_dual_path_tower}
\end{equation}
The trained predictor $f_0$ is $\mathcal A$-measurable, and $f_\eta$ is $\mathcal B$-measurable. For any square-integrable $\mathcal B$-measurable noise predictor $h$, whose output is determined by $(U,S,G)$, conditional orthogonality gives
\begin{equation}
\mathbb E\langle\epsilon-m_{SG},h-m_{SG}\rangle=0.
\label{eq:prostyle_dual_path_bayes_orthogonality}
\end{equation}
Expanding $\epsilon-h=(\epsilon-m_{SG})-(h-m_{SG})$ therefore yields
\begin{equation}
\mathcal R(h)=\mathcal R(m_{SG})+
\mathbb E\|h-m_{SG}\|_2^2.
\label{eq:prostyle_dual_path_risk_projection}
\end{equation}
Apply this identity to $f_0$ and $f_\eta$, and subtract.
\begin{equation}
\begin{aligned}
\mathcal R(f_0)-\mathcal R(f_\eta)
={}&\mathbb E\|m_{SG}-f_0\|_2^2\\
&-\mathbb E\|f_\eta-m_{SG}\|_2^2.
\end{aligned}
\label{eq:prostyle_dual_path_risk_difference}
\end{equation}
Next, decompose $m_{SG}-f_0=(m_{SG}-m_S)+(m_S-f_0)$. The second term is $\mathcal A$-measurable, so Eq.~\eqref{eq:prostyle_dual_path_tower} makes the expected cross term zero. Consequently,
\begin{equation}
\begin{aligned}
\mathbb E\|m_{SG}-f_0\|_2^2
={}&\mathbb E\|m_{SG}-m_S\|_2^2\\
&+\mathbb E\|m_S-f_0\|_2^2.
\end{aligned}
\label{eq:prostyle_dual_path_correctable_error}
\end{equation}
Substitution into Eq.~\eqref{eq:prostyle_dual_path_risk_difference} proves Proposition~\ref{thm:prostyle_dual_path_gain}.

\paragraph{Connection to native residual learning.}
Define the realized complete-denoiser correction $\delta_\eta$ and the optimal predictable correction $\delta^\star$ by
\begin{equation}
\delta_\eta=f_\eta-f_0,\qquad
\delta^\star=m_{SG}-f_0.
\label{eq:prostyle_dual_path_corrections}
\end{equation}
The difference $\delta_\eta$ is measured at the output of the complete denoiser. It incorporates all downstream effects of native injection, including changes in subsequent style-conditioned attention. This definition applies directly to the nonlinear full-network predictor.

Because $f_0$ is $\mathcal B$-measurable, the optimal correction also satisfies
\begin{equation}
\delta^\star=\mathbb E[\epsilon-f_0\mid U,S,G].
\label{eq:prostyle_dual_path_target_residual}
\end{equation}
Using Eq.~\eqref{eq:prostyle_dual_path_risk_projection},
\begin{equation}
\mathcal R(f_\eta)=\mathcal R(m_{SG})+
\mathbb E\|\delta_\eta-\delta^\star\|_2^2.
\label{eq:prostyle_dual_path_residual_objective}
\end{equation}
For fixed representations and $f_0$, the first term is independent of the native attention parameters. Hence minimizing the population noise-prediction objective in Eq.~\eqref{eq:cross_content_loss} over those parameters is equivalent to approximating $\delta^\star$. This decomposes the existing objective into the minimum risk associated with the dual-path inputs and a residual approximation error. The first term in Proposition~\ref{thm:prostyle_dual_path_gain} isolates information supplied by $G$ beyond $S$, while the second retains approximation error in the trained predictor $f_0$.

\paragraph{Zero initialization and inclusion of the baseline.}
Let $\eta_0$ denote a native-pathway parameter configuration with well-defined internal operations and all output projections, including any biases, equal to zero. At each primary layer, the native update in Eq.~\eqref{eq:dual_pathway} then vanishes. The auxiliary layers retain the style-only updates. Induction through the denoiser therefore gives identical hidden states and final predictions to the style-only model.
\begin{equation}
f_{\eta_0}=f_0,\qquad
\delta_{\eta_0}=0,\qquad
\mathcal R(f_{\eta_0})=\mathcal R(f_0).
\label{eq:prostyle_dual_path_zero_initialization}
\end{equation}
Thus, for any set $\Theta_n$ of admissible native-pathway parameter configurations containing $\eta_0$,
\begin{equation}
\inf_{\eta\in\Theta_n}\mathcal R(f_\eta)
\le\mathcal R(f_0).
\label{eq:prostyle_dual_path_baseline_inclusion}
\end{equation}
Here $\inf$ denotes the greatest lower bound over the indicated parameter set. Zero initialization places the trained baseline within the dual-path function class.
\end{proof}

The Stage-2 analysis concerns population denoising risk with the reference representations and style pathway fixed. Conditional mean independence is an assumption motivated by cross-content pairing. The risk bounds characterize attainable predictors, while finite-data optimization, perceptual outcomes, and Stage-3 behavior are evaluated empirically.

\section{Implementation Details}
\label{sec:supp_implementation}

\begin{table*}[t]
	\centering
	\caption{
		Training configurations of ProStyle.
	}
	\label{tab:supp_training_configuration}
	\small
	\begin{tabular}{lccc}
		\toprule
		Configuration & Stage 1 & Stage 2 & Stage 3 \\
		\midrule
		Training updates & 40,000 & 8,000 & 7,000 \\
		Resampler learning rate & $10^{-4}$ & Frozen & $10^{-5}$ \\
		Style-side K/V learning rate & $10^{-4}$ & Frozen & $2\times10^{-5}$ \\
		Native pathway learning rate & -- & $10^{-4}$ & $10^{-4}$ \\
		Trainable parameters & 58,374,144 & 16,650,240 & 75,024,384 \\
		Batch size & 8 & 8 & 8 \\
		Mixed precision & BF16 & BF16 & BF16 \\
		\bottomrule
	\end{tabular}
\end{table*}

\begin{table}[t]
	\centering
	\caption{Effect of DDIM sampling steps on ReConStyle-Full. Time reports the separately measured mean per-image sampling latency. Bold indicates the best result in each column.}
	\label{tab:supp_ddim}
	\small
	\setlength{\tabcolsep}{4pt}
	\begin{tabular}{lccccc}
		\toprule
		Steps & Style-CSD$\uparrow$ & CLIP-T$\uparrow$ & TCF$\uparrow$ & ROO$\downarrow$ & Time (s)$\downarrow$ \\
		\midrule
		5   & 0.3381 & 0.2584 & 0.6550 & \textbf{0.0212} & \textbf{0.33} \\
		20  & 0.4057 & \textbf{0.2680} & 0.7641 & 0.0341 & 0.97 \\
		50  & \textbf{0.4076} & 0.2668 & \textbf{0.7729} & 0.0246 & 2.15 \\
		200 & 0.4068 & 0.2662 & 0.7688 & 0.0325 & 8.53 \\
		\bottomrule
	\end{tabular}
\end{table}

\begin{table*}[t]
\centering
\caption{
Inference configurations of the evaluated methods. For SPG, reference scale denotes the style-posterior guidance scale.
}
\label{tab:supp_baseline_configuration}
\small
\setlength{\tabcolsep}{5pt}
\begin{tabular}{llcccc}
\toprule
Method & Backbone & Sampler & Steps & CFG & Ref. scale \\
\midrule
IP-Adapter & SD3.5-Large & Flow-matching Euler & 24 & 5.0 & 1.0 \\
\midrule
InstantStyle & SDXL-Base-1.0 & Euler & 30 & 5.0 & 1.0 \\
StyleStudio & SDXL-Base-1.0 & Euler & 50 & 5.0 & 1.0 \\
SPG & SDXL-Base-1.0 & DDIM & 50 & 7.0 & 3.0 \\
\midrule
DEADiff & SD1.5 & DDIM & 50 & 8.0 & 1.0 \\
StyleShot & SD1.5 & DDIM & 50 & 8.0 & 1.0 \\
Less is More & SD1.5 & DDIM & 50 & 8.0 & 1.0 \\
ProStyle & SD1.5 & DDIM & 50 & 8.0 & 1.0 \\
\bottomrule
\end{tabular}
\end{table*}

\subsection{ProStyle Architecture and Training}
\label{sec:supp_prostyle_training}

\begin{table*}[t]
\centering
\caption{
Learned conditioning parameters and per-image inference latency on an NVIDIA RTX 5880 Ada GPU. Parameter counts exclude frozen backbones and encoders. Latency is reported as mean $\pm$ sample standard deviation over five measurements. AMP denotes automatic mixed precision.
}
\label{tab:supp_computation}
\small
\begin{tabular}{lrlc}
\toprule
Method & Learned conditioning (M) & Precision & Time (s/image) \\
\midrule
IP-Adapter & 797.926 & BF16 & $21.149\pm0.243$ \\
InstantStyle & 351.285 & FP16 & $5.002\pm0.035$ \\
StyleStudio & 507.896 & FP16 & $20.252\pm0.128$ \\
SPG & 0.000 & BF16 & $16.645\pm0.155$ \\
DEADiff & 427.215 & FP16 AMP & $1.891\pm0.008$ \\
StyleShot & 159.610 & FP32 & $3.497\pm0.032$ \\
Less is More & 159.610 & FP32 & $3.528\pm0.011$ \\
\midrule
\textbf{ProStyle} & 75.024 & FP16 AMP & $2.151\pm0.016$ \\
\bottomrule
\end{tabular}
\end{table*}

\paragraph{Architecture.}
ProStyle augments Stable Diffusion v1.5 with two pathways that share a frozen CLIP ViT-L/14 image encoder. Reference images are converted to RGB, resized to $224\times224$ with antialiased bicubic interpolation, and normalized using the CLIP channel statistics. We extract the last CLIP hidden layer, discard the class token, and retain 256 patch tokens with 1,024 dimensions each. The style pathway compresses these features through a four-block Resampler. It uses a hidden dimension of 768, 12 attention heads, 16 learned queries, and a feed-forward expansion ratio of four. Each block contains latent self-attention, cross-attention to the image features, and a feed-forward network. The resulting 16 style tokens have 768 dimensions and feed independently learned style-side key/value projections.

The native reference pathway normalizes the CLIP patch features with layer normalization. At each primary layer, it uses independent reference key/value and output projections. The query projection is shared with the pretrained attention layer and remains frozen. Reference key/value projections use Xavier initialization, and output projections have no bias and are zero-initialized. The native residual scale is fixed at $\lambda_n=1$.

The style pathway conditions all 16 cross-attention layers. Its scale is 1.0 at the 11 primary layers and 0.05 at the five auxiliary layers. In execution order, the primary indices are $\{0,1,2,3,9,10,11,12,13,14,15\}$, and the auxiliary indices are $\{4,5,6,7,8\}$. Native reference attention is applied only at primary layers. The U-Net backbone, VAE, text encoder, and CLIP image encoder remain frozen throughout training.

\paragraph{Training.}
All three stages use the same 134,000 cross-content training pairs and content-only target descriptions from Appendix~\ref{sec:supp_benchmark}. Reference and target assignments remain fixed. The diffusion noise-prediction objective is shared across stages, with no additional style-consistency or reconstruction loss. Training uses AdamW with $\beta_1=0.9$, $\beta_2=0.999$, $\epsilon=10^{-8}$, weight decay 0.01, and gradient-norm clipping at 1.0. We use a batch size of 8, one gradient-accumulation step, and BF16 mixed precision on one NVIDIA RTX 5880 Ada GPU.

Stage 1 trains the Resampler and style-side key/value projections for 40,000 updates. Stage 2 starts from this checkpoint, freezes the style pathway, and trains the native pathway for 8,000 updates. Stage 3 then jointly refines both pathways for 7,000 updates. This schedule first establishes the style pathway, then fits a native residual, and finally coordinates their parameters. Stage 1 and Stage 2 use a 500-update linear learning-rate warmup followed by a constant rate. Parameters unfrozen in Stage 3 start at the rates in Table~\ref{tab:supp_training_configuration}, without another warmup. No conditioning dropout is used. ProStyle results use the final Stage-3 checkpoint after 55,000 total updates.

\subsection{Inference Settings}
\label{sec:supp_inference}

\paragraph{ProStyle Inference.}
ProStyle uses 50-step DDIM sampling, classifier-free guidance scale 8.0, image-conditioning scale 1.0, and $\eta=0$. The unconditional branch uses the negative prompt with zeroed style tokens. Both pathways use features from one frozen image-encoder forward pass. Reference encoding and token preparation occur before sampling. Denoising and VAE decoding use FP16 automatic mixed precision, while training uses BF16 as specified above. Primary-layer, auxiliary-layer, and native conditioning scales match their training values.

\paragraph{Inference Seeds.}
All quantitative evaluations on ReConStyle-Full use five inference seed configurations, while those on ReConStyle-Rapid use three. Metrics are computed separately for each configuration and then averaged across configurations.

\paragraph{Number of Sampling Steps.}
\label{sec:supp_ddim_steps}
We evaluate 5, 20, 50, and 200 DDIM steps on ReConStyle-Full. Table~\ref{tab:supp_ddim} shows that five steps reduce sampling time but yield substantially lower style similarity and target-content fidelity. Increasing the sampling budget from 20 to 50 steps improves TCF and lowers ROO while maintaining similar Style-CSD. Further increasing to 200 steps provides no improvement in these three metrics and takes approximately four times as long as 50 steps. We therefore use 50 steps for the main evaluations.

\subsection{Baseline Implementation Details}
\label{sec:supp_baselines}

We use released baseline weights with their corresponding implementation settings. Methods receive the same reference images, target prompts, and numerical seeds. Each method follows its released preprocessing and backbone configuration. Table~\ref{tab:supp_baseline_configuration} summarizes the sampling settings. IP-Adapter uses the InstantX SD3.5-Large implementation and released weights~\cite{instantx2024sd35ipadapter}. InstantStyle, StyleStudio, and SPG use the SDXL backbone~\cite{podell2024sdxl}. Because reference-conditioning scales act on different quantities across methods, each method uses its released default scale.

\paragraph{Computational Cost.}
Table~\ref{tab:supp_computation} reports the learned conditioning-module parameter counts and single-image inference latency. Parameter counts exclude frozen generative backbones and pretrained encoders. Latency is measured on one NVIDIA RTX 5880 Ada GPU using a common reference--prompt condition. Each method performs two warmup calls followed by five timed calls, with CUDA synchronization before and after each generation. Timing includes reference preprocessing, conditioning, denoising, and decoding, while excluding model initialization, image serialization, and metric computation. Latency is measured under each method's evaluated inference configuration. Among the evaluated configurations, ProStyle ranks among the fastest methods and uses the fewest learned conditioning parameters among training-based methods.

\section{User Study}
\label{sec:supp_user_study}

\begin{figure*}[t]
	\centering
	\includegraphics[width=\textwidth]{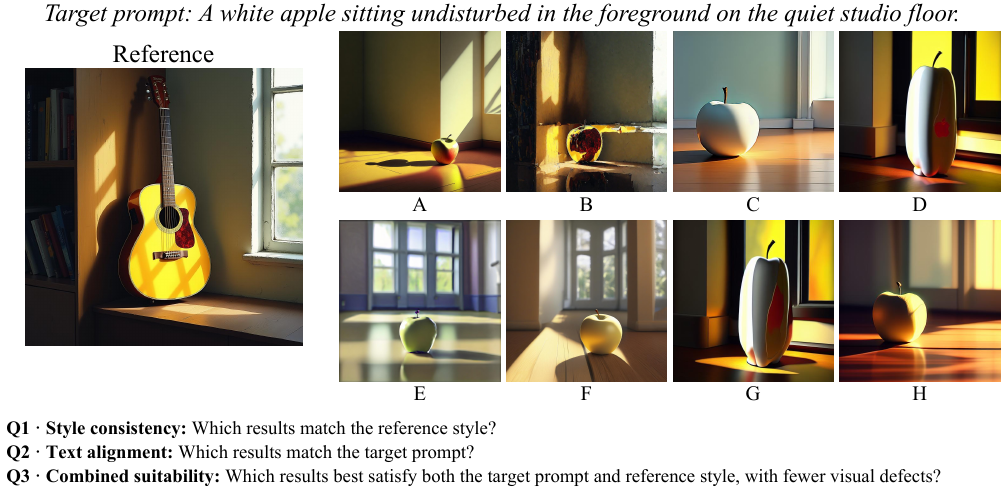}
	\caption{Example user-study case with eight anonymous outputs. Q1 (style consistency) and Q2 (text alignment) each allow 0--8 selections, while Q3 (combined suitability) requires 1--2.}
	\label{fig:supp_user_study_interface}
\end{figure*}

\subsection{Evaluation Task and Interface}
\label{sec:supp_user_study_protocol}

The study evaluates reference-style consistency, text alignment, and combined suitability as separate judgments. It includes 26 participants with education ranging from junior college to the master's level and majors including accounting, computer science, and mechanical engineering. Participation was voluntary and compensated. All participants gave informed consent and received a brief introduction to the task background before the study. We randomly sample 30 cases from the ReConStyle-Full test set. The target subjects differ from the reference content. All participants evaluate the same cases, yielding 780 evaluations across participants and cases.

Each case displays the reference, target prompt, and outputs from ProStyle and seven baselines. Anonymous labels A--H conceal method names, and their mapping changes across cases. Images can be enlarged for detail inspection. All outputs are resized to $512\times512$ before display. Figure~\ref{fig:supp_user_study_interface} illustrates one case.

\paragraph{Style Consistency.}
Participants select all outputs that match the reference style. They consider texture, brushwork, material, and overall visual treatment. 
Zero to eight selections are permitted.

\paragraph{Text Alignment.}
Participants check the subject, attributes, count, action, and scene against the target prompt. Zero to eight selections are permitted.

\paragraph{Combined Suitability.}
Participants select the one or two most suitable outputs by jointly considering prompt adherence, reference-style similarity, and visible defects. This criterion is evaluated separately from the style and text selections.

\subsection{Selection Counts and Aggregation}
\label{sec:supp_user_study_aggregation}

Table~\ref{tab:supp_user_study_counts} reports selections for each method and dimension after mapping the case-specific A--H labels back to methods.

\begin{table}[t]
	\centering
	\caption{Selection counts out of 780 evaluations. Bold indicates the highest count in each dimension.}
	\label{tab:supp_user_study_counts}
	\small
	\setlength{\tabcolsep}{4pt}
	\begin{tabular}{lccc}
		\toprule
		Method & Style & Text & Suitability \\
		\midrule
		IP-Adapter & \textbf{394} & 135 & 84 \\
		InstantStyle & 288 & 288 & 116 \\
		StyleStudio & 137 & 419 & 112 \\
		SPG & 273 & 72 & 29 \\
		DEADiff & 202 & 389 & 95 \\
		StyleShot & 315 & 193 & 64 \\
		Less is More & 323 & 228 & 69 \\
		\midrule
		ProStyle & 330 & \textbf{658} & \textbf{397} \\
		\bottomrule
	\end{tabular}
\end{table}

The selection rate is the fraction of the 780 evaluations in which a method is selected for a dimension. Let $N_{m,d}$ be the count for method $m$ and dimension $d$.
\begin{equation}
P_{m,d}=\frac{N_{m,d}}{26\times30}\times100\%.
\label{eq:user_study_selection_rate}
\end{equation}
For example, ProStyle's 658 text-alignment selections correspond to $658/780=84.36\%$. Multiple methods can be selected, so column percentages need not sum to 100\%.

\subsection{Analysis of User Preferences}
\label{sec:supp_user_study_analysis}

ProStyle leads in text alignment and combined suitability, exceeding StyleStudio and InstantStyle by 30.64 and 36.03 percentage points, respectively, while ranking second in style at 42.31\%. IP-Adapter receives the most style selections at 50.51\% but far fewer text and suitability selections, suggesting that its perceived style advantage does not translate into prompt adherence or overall suitability.

\subsection{Completion Time and Response Coverage}
\label{sec:supp_user_study_time}

The mean completion time is 44.2 minutes and the median is 40.7 minutes for 30 cases, approximately 1.5 minutes per case. During this time, participants inspect the reference, prompt, and eight outputs and make three separate judgments for each case. All 26 questionnaires cover all 30 cases, providing 780 complete evaluations and equal opportunities for each method to be selected.

\section{Detailed Experimental Results}
\label{sec:supp_additional_results}

\begin{figure*}[t]
	\centering
	\includegraphics[width=\textwidth]{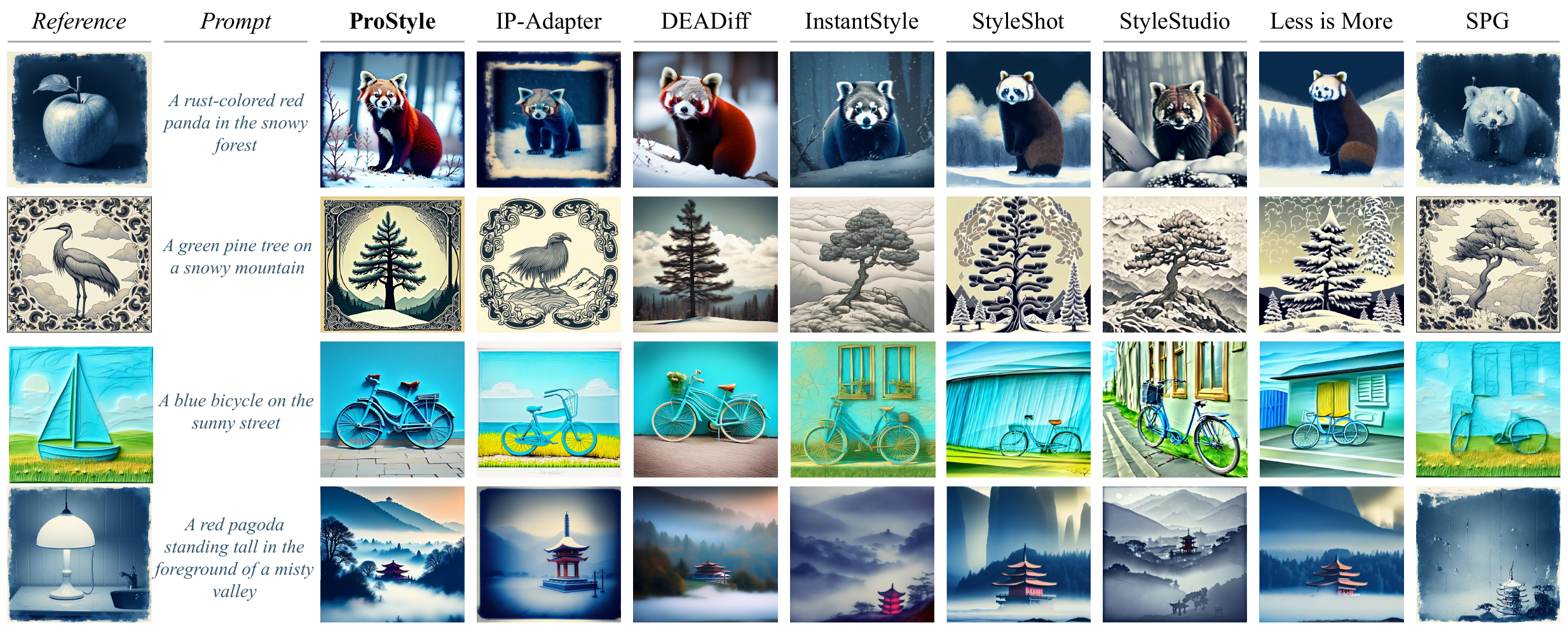}
	\caption{
		Additional comparisons on ReConStyle-Full examine transfer across different styles and target subjects.
		All methods are evaluated using the same reference images and target prompts.
	}
	\label{fig:supp_D_comparisons}
\end{figure*}

\begin{table*}[t]
	\centering
	\caption{Comparison on the seen and unseen style subsets of ReConStyle-Full. Bold indicates the best result for each metric within each subset.}
	\label{tab:supp_seen_unseen}
	\small
	\setlength{\tabcolsep}{3.5pt}
	\begin{tabular}{lcccccccc}
		\toprule
		& \multicolumn{4}{c}{Seen styles (40)} & \multicolumn{4}{c}{Unseen styles (10)} \\
		\cmidrule(lr){2-5}\cmidrule(lr){6-9}
		Method & Style-CSD$\uparrow$ & CLIP-T$\uparrow$ & TCF$\uparrow$ & ROO$\downarrow$
		& Style-CSD$\uparrow$ & CLIP-T$\uparrow$ & TCF$\uparrow$ & ROO$\downarrow$ \\
		\midrule
		IP-Adapter & 0.4024 & 0.2183 & 0.6234 & 0.1740 & \textbf{0.4409} & 0.2055 & 0.5702 & 0.1943 \\
		InstantStyle & 0.4026 & 0.2555 & 0.7045 & 0.0449 & 0.4355 & 0.2545 & 0.7032 & 0.0515 \\
		StyleStudio & 0.3133 & \textbf{0.2682} & \textbf{0.7926} & \textbf{0.0227} & 0.3559 & 0.2661 & \textbf{0.7811} & 0.0272 \\
		SPG & 0.3464 & 0.1947 & 0.4171 & 0.0895 & 0.3889 & 0.1918 & 0.4297 & 0.0893 \\
		DEADiff & 0.3657 & 0.2580 & 0.6640 & 0.0277 & 0.4121 & 0.2504 & 0.6418 & 0.0339 \\
		StyleShot & 0.3768 & 0.2351 & 0.6484 & 0.0464 & 0.4132 & 0.2301 & 0.6539 & 0.0602 \\
		Less is More & 0.3578 & 0.2409 & 0.6597 & 0.0394 & 0.3914 & 0.2360 & 0.6645 & 0.0493 \\
		\midrule
		ProStyle & \textbf{0.4105} & 0.2668 & 0.7747 & 0.0242 & 0.3957 & \textbf{0.2667} & 0.7658 & \textbf{0.0265} \\
		\bottomrule
	\end{tabular}
\end{table*}

\begin{table*}[t]
	\centering
	\caption{Results on Style-Rank with 432 conditions per method. CLIP-I and IR denote CLIP-Image and ImageReward. Bold indicates the best value. \textbf{*}\,Similarity to the reference image, which rises with both style similarity and reference-content leakage.}
	\label{tab:supp_style_rank}
	\small
	\setlength{\tabcolsep}{8pt}
	\begin{tabular}{lccccc}
		\toprule
		Method & CLIP-Text$\uparrow$ & CLIP-I\textbf{*}$\uparrow$ & DINOv2\textbf{*}$\uparrow$ & IR$\uparrow$ & ROO$\downarrow$ \\
		\midrule
		IP-Adapter   & 18.718 & \textbf{75.348} & \textbf{36.937} & -1.276 & 0.3750 \\
		InstantStyle & 22.094 & 64.767 & 18.804 & -0.278 & 0.0556 \\
		StyleStudio  & 20.853 & 56.939 & 14.943 & -1.223 & \textbf{0.0060} \\
		SPG          & 21.337 & 68.457 & 28.239 & -1.204 & 0.0770 \\
		DEADiff      & 22.932 & 60.919 & 18.481 &  0.073 & 0.0437 \\
		StyleShot    & 21.994 & 63.307 & 22.473 & -0.333 & 0.0685 \\
		Less is More & 22.391 & 62.351 & 20.695 & -0.264 & 0.0426 \\
		\midrule
		ProStyle     & \textbf{23.463} & 64.512 & 23.601 & \textbf{0.214} & 0.0161 \\
		\bottomrule
	\end{tabular}
\end{table*}

\subsection{Results on Seen and Unseen Styles}
\label{sec:supp_style_generalization}

To examine transfer beyond training styles, we separate the 40 seen styles and 10 unseen styles. The seen and unseen subsets contain 5,280 and 1,320 of the 6,600 Full conditions. Table~\ref{tab:supp_seen_unseen} compares methods within each subset, because their reference styles and target conditions differ.

On seen styles, ProStyle has the highest category-level Style-CSD of 0.4105, with content-related scores close to the best values. On unseen styles, it leads on CLIP-T and ROO and ranks second on TCF, supporting transfer of content control within these 10 unseen styles. On unseen styles, IP-Adapter, InstantStyle, StyleShot, and DEADiff obtain higher Style-CSD than ProStyle, while ProStyle achieves lower ROO and higher TCF than all four.

\begin{table*}[t]
	\centering
	\caption{Reference-conditioning strength sweep on ReConStyle-Rapid. For SPG, the style-posterior guidance scale is $3\alpha$.}
	\label{tab:supp_strength_sweep}
	\small
	\setlength{\tabcolsep}{6pt}
	\begin{tabular}{lcccccccc}
		\toprule
		Metric & $\alpha$ & IP-Adapter & InstantStyle & SPG & DEADiff & StyleShot & Less is More & ProStyle \\
		\midrule
		Style-CSD$\uparrow$ & 0.5 & 0.3630 & 0.3666 & 0.3528 & 0.1067 & 0.3746 & 0.3300 & 0.2684 \\
		& 0.75 & 0.4058 & 0.4160 & 0.3827 & 0.2480 & 0.4206 & 0.3933 & 0.3616 \\
		& 1.0 & 0.4168 & 0.4315 & 0.3965 & 0.4177 & 0.4294 & 0.4072 & 0.4237 \\
		& 1.25 & 0.4099 & 0.4338 & 0.4059 & 0.4214 & 0.4303 & 0.4099 & 0.4234 \\
		\midrule
		CLIP-T$\uparrow$ & 0.5 & 0.2593 & 0.2590 & 0.2311 & 0.2633 & 0.2553 & 0.2583 & 0.2668 \\
		& 0.75 & 0.2470 & 0.2514 & 0.2188 & 0.2622 & 0.2458 & 0.2495 & 0.2598 \\
		& 1.0 & 0.2314 & 0.2431 & 0.2122 & 0.2468 & 0.2378 & 0.2410 & 0.2475 \\
		& 1.25 & 0.2114 & 0.2371 & 0.2077 & 0.2274 & 0.2317 & 0.2352 & 0.2397 \\
		\midrule
		TCF$\uparrow$ & 0.5 & 0.8331 & 0.8308 & 0.6403 & 0.8056 & 0.8142 & 0.8128 & 0.7933 \\
		& 0.75 & 0.7603 & 0.7914 & 0.5725 & 0.7786 & 0.7719 & 0.7814 & 0.7900 \\
		& 1.0 & 0.6175 & 0.7453 & 0.5203 & 0.7181 & 0.7264 & 0.7283 & 0.7917 \\
		& 1.25 & 0.4703 & 0.6961 & 0.4853 & 0.5689 & 0.6967 & 0.6908 & 0.7642 \\
		\midrule
		ROO$\downarrow$ & 0.5 & 0.1304 & 0.1119 & 0.1220 & 0.0889 & 0.1152 & 0.1061 & 0.1038 \\
		& 0.75 & 0.1615 & 0.1333 & 0.1561 & 0.1075 & 0.1303 & 0.1301 & 0.1104 \\
		& 1.0 & 0.2439 & 0.1399 & 0.1781 & 0.1247 & 0.1505 & 0.1393 & 0.1094 \\
		& 1.25 & 0.3242 & 0.1437 & 0.2160 & 0.1415 & 0.1495 & 0.1282 & 0.1329 \\
		\bottomrule
	\end{tabular}
\end{table*}

\subsection{Results on Style-Rank}
\label{sec:supp_style_rank}
We additionally evaluate ProStyle and seven baselines on Style-Rank~\cite{benaroche2024stylerank}, using its 108 reference images and publicly released prompt pool. We randomly sample four distinct prompts per reference and fix them for all methods, yielding 432 conditions per method, while retaining each method's inference settings. Results are averaged over five inference seed configurations. Following the Style-Rank implementation, we report mean CLIP-Text, CLIP-Image~\cite{radford2021learning}, DINOv2~\cite{DBLP:journals/tmlr/OquabDMVSKFHMEA24}, and ImageReward~\cite{xu2023imagereward} scores. The CLIP and DINOv2 similarity scores are multiplied by 100, while ImageReward retains its original scale.
Table~\ref{tab:supp_style_rank} reports the results. ProStyle achieves the highest CLIP-Text and ImageReward scores of 23.463 and 0.214, respectively. These results support its ability to follow target prompts and produce outputs with higher predicted human preference across the evaluated conditions. CLIP-Image and DINOv2 measure similarity between generated and reference images, which can reflect shared content as well as stylistic appearance. IP-Adapter obtains the highest scores on these two metrics together with the lowest CLIP-Text score, a pattern consistent with reference-image similarity being influenced by content correspondence. We compute ROO on the same outputs using the rules in Appendix~\ref{sec:supp_roo}, with the subject of each prompt as the target. IP-Adapter and SPG have the highest ROO of 0.3750 and 0.0770, and ProStyle has the second-lowest ROO of 0.0161. Together, these evaluations provide complementary assessments of prompt alignment, predicted human preference, and reference-image similarity on an external benchmark.

\subsection{Reference-Conditioning Strength Sweep}
\label{sec:supp_strength_sweep}

This experiment characterizes how reference-conditioning strength changes category-level style similarity and content control. We sweep $\alpha\in\{0.5,0.75,1.0,1.25\}$ for ProStyle and six baselines on all 300 ReConStyle-Rapid conditions. The default is $\alpha=1.0$, and SPG uses style-posterior guidance $3\alpha$. For ProStyle, $\alpha$ multiplies the layer weights $\alpha_l$ in Eq.~\eqref{eq:asymmetric_injection}, scaling primary and auxiliary layers proportionally. Other settings follow Table~\ref{tab:supp_baseline_configuration}. We exclude StyleStudio from the sweep because changing its reference scale does not affect its outputs in the evaluated configuration.

Table~\ref{tab:supp_strength_sweep} reports the complete results, and Figure~\ref{fig:strength_curves} visualizes the trade-offs between Style-CSD and CLIP-T, TCF, and ROO. Across the baselines, weaker conditioning generally preserves content control at the cost of style similarity, while stronger conditioning improves style but reduces prompt adherence and increases reference-object overlap. For example, IP-Adapter shows a pronounced decline in TCF and increase in ROO as conditioning becomes stronger, whereas DEADiff gains substantial style similarity but also loses target-content fidelity. These results show that adjusting strength only shifts the balance between style and content. By contrast, increasing $\alpha$ from 0.5 to 1.0 raises ProStyle's Style-CSD from 0.2684 to 0.4237 while maintaining TCF between 0.7900 and 0.7933, demonstrating stronger stylistic expression with stable target-subject fidelity over this range. At the default strength, ProStyle achieves style similarity close to the strongest baselines, together with higher CLIP-T and TCF and lower ROO than all six baselines.

\begin{table*}[t]
	\centering
	\caption{
		Component and optimization ablations on ReConStyle-Full. The first four rows examine conditioning and pathway configurations, while the last four examine continued training and dual-pathway optimization. Results are reported as mean $\pm$ sample standard deviation across five inference seed configurations. Bold indicates the best mean in each column.
	}
	\label{tab:supp_component_ablation}
	\small
	\setlength{\tabcolsep}{4pt}
	\begin{tabular}{lcccc}
		\toprule
		Variant & Style-CSD$\uparrow$ & CLIP-T$\uparrow$ & TCF$\uparrow$ & ROO$\downarrow$ \\
		\midrule
		Style pathway, zero auxiliary scale
		& $0.3529 \pm 0.0033$ & $0.2657 \pm 0.0011$ & $0.7733 \pm 0.0051$ & $\mathbf{0.0215 \pm 0.0015}$ \\
		Style pathway, strong auxiliary conditioning
		& $0.3866 \pm 0.0035$ & $0.2591 \pm 0.0018$ & $0.7616 \pm 0.0077$ & $0.0298 \pm 0.0045$ \\
		Style pathway, weak auxiliary conditioning
		& $0.3811 \pm 0.0059$ & $0.2645 \pm 0.0017$ & $0.7746 \pm 0.0044$ & $0.0250 \pm 0.0022$ \\
		Native pathway only
		& $0.3863 \pm 0.0076$ & $0.2668 \pm 0.0016$ & $0.7551 \pm 0.0046$ & $0.0294 \pm 0.0055$ \\
		\midrule
		Style pathway continuation
		& $0.3731 \pm 0.0057$ & $0.2656 \pm 0.0015$ & $0.7693 \pm 0.0044$ & $0.0312 \pm 0.0052$ \\
		Dual pathways, joint optimization
		& $0.3996 \pm 0.0057$ & $\mathbf{0.2669 \pm 0.0019}$ & $\mathbf{0.7756 \pm 0.0065}$ & $0.0251 \pm 0.0028$ \\
		Dual pathways, extended Stage 2
		& $0.3998 \pm 0.0062$ & $\mathbf{0.2669 \pm 0.0019}$ & $0.7667 \pm 0.0075$ & $0.0318 \pm 0.0046$ \\
		\textbf{ProStyle}
		& $\mathbf{0.4076 \pm 0.0055}$ & $0.2668 \pm 0.0020$ & $0.7729 \pm 0.0078$ & $0.0246 \pm 0.0020$ \\
		\bottomrule
	\end{tabular}
\end{table*}
\subsection{Transfer to Other Backbones}
\label{sec:supp_backbone_transfer}

\begin{table*}[t]
	\centering
	\caption{ProStyle and baselines grouped by backbone on ReConStyle-Rapid at the default reference strength. Bold indicates the best value within each backbone.}
	\label{tab:supp_backbone}
	\small
	\setlength{\tabcolsep}{10pt}
	\begin{tabular}{llcccc}
		\toprule
		Backbone & Method & Style-CSD$\uparrow$ & CLIP-T$\uparrow$ & TCF$\uparrow$ & ROO$\downarrow$ \\
		\midrule
		SD1.5 & DEADiff & 0.4177 & 0.2468 & 0.7181 & 0.1247 \\
		& StyleShot & \textbf{0.4294} & 0.2378 & 0.7264 & 0.1505 \\
		& Less is More & 0.4072 & 0.2410 & 0.7283 & 0.1393 \\
		& ProStyle & 0.4237 & \textbf{0.2475} & \textbf{0.7917} & \textbf{0.1094} \\
		\midrule
		SD2.1 & ProStyle & 0.4010 & 0.2586 & 0.7817 & 0.0990 \\
		\midrule
		SDXL & InstantStyle & \textbf{0.4315} & 0.2431 & 0.7453 & 0.1399 \\
		& SPG & 0.3965 & 0.2122 & 0.5203 & 0.1781 \\
		& ProStyle & 0.3889 & \textbf{0.2570} & \textbf{0.8117} & \textbf{0.0704} \\
		\midrule
		SD3.5-L & IP-Adapter & \textbf{0.4168} & 0.2314 & 0.6175 & 0.2439 \\
		& ProStyle & 0.3864 & \textbf{0.2671} & \textbf{0.8178} & \textbf{0.0696} \\
		\bottomrule
	\end{tabular}
\end{table*}

We apply ProStyle to SD2.1, SDXL, and SD3.5-Large and evaluate it on ReConStyle-Rapid. SD2.1 has the same 16 cross-attention layers as SD1.5, so we keep the primary and auxiliary layers of Section~\ref{sec:hierarchical_asymmetric_reference_injection} and the auxiliary scale of 0.05. SDXL has 70 cross-attention layers, and we use the 20 layers of its lowest-resolution down-sampling stage as primary layers and apply weak conditioning with a scale of 0.01 to the remaining 50 layers. On both U-Net backbones, the style keys and values are concatenated with the text keys and values, and the native pathway adds its residual at the primary layers, as in Section~\ref{sec:method}. SD3.5-Large replaces cross-attention with joint attention over image and text tokens. We rank its 38 joint-attention blocks by style relevance following Conditional Balance~\cite{cohen2025conditional}, select the top 28 as primary blocks, append the style keys and values to the image-token attention of these blocks, and add the native residual at the same blocks.

Table~\ref{tab:supp_backbone} shows that all three backbones raise CLIP-T over SD1.5, and SDXL and SD3.5-Large also reach higher TCF and lower ROO. On SDXL and SD3.5-Large, ProStyle also achieves higher CLIP-T and TCF and lower ROO than the baselines built on the same backbone. This is a preliminary small-scale transfer, in which the layer grouping and injection design follow SD1.5 without studying the injection characteristics of each backbone. In future work, we will analyze how individual layers of each backbone encode style and content, and use this analysis to decide where and how strongly to inject the style tokens.

\subsection{Additional Qualitative Comparisons}
\label{sec:supp_additional_qualitative}

Figure~\ref{fig:supp_D_comparisons} presents selected cases with different reference and target subjects. ProStyle renders the requested rust-colored red panda, while StyleShot and Less is More show black-and-white panda-like markings and IP-Adapter and InstantStyle miss the rust color. For the pine tree, IP-Adapter reproduces the reference bird and omits the tree. ProStyle retains the decorative framing while depicting the requested tree. For the bicycle, ProStyle places the requested blue bicycle on a paved street with the reference palette and simplified patterns, whereas IP-Adapter sets it in a field. Its red pagoda retains the misty atmosphere, while some baselines weaken or omit the requested structure. These ProStyle outputs combine reference appearance cues with the requested target subjects.

\begin{figure}[t]
	\centering
	\includegraphics[width=\linewidth]{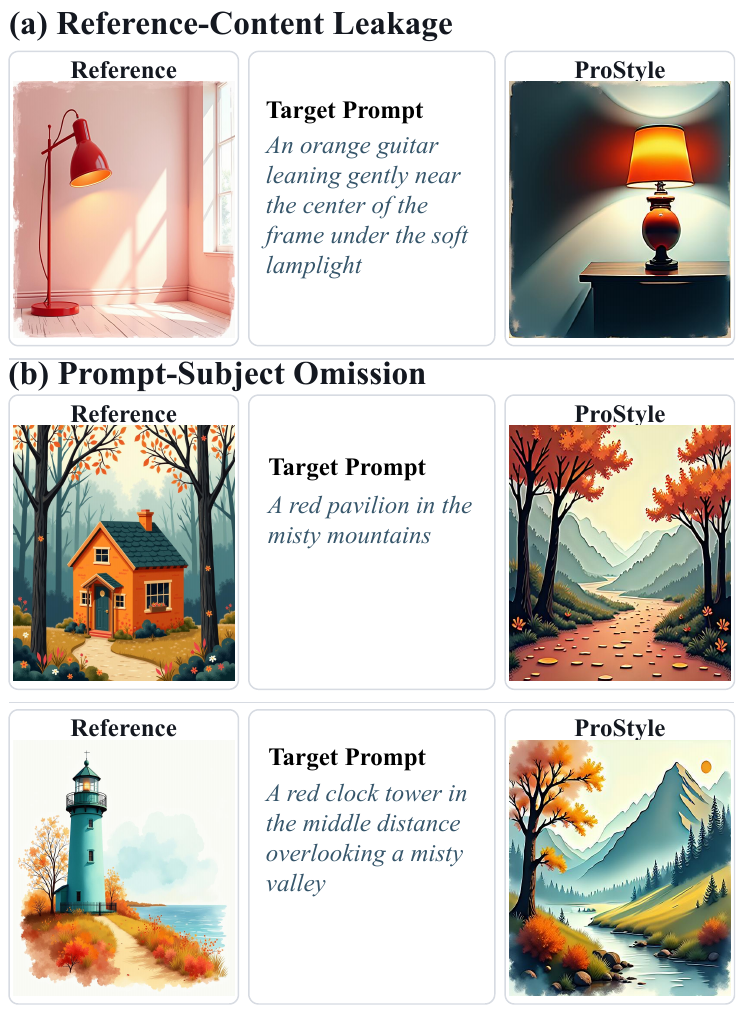}
	\caption{
		Representative failure cases of ProStyle.
		(a) Reference-content leakage.
		(b) Prompt-subject omission.
	}
	\label{fig:supp_D_failures}
\end{figure}
\subsection{Failure Cases}
\label{sec:supp_failure_cases}

Figure~\ref{fig:supp_D_failures} illustrates two types of content-control failure. In (a), ProStyle generates a lamp, matching the reference object category, instead of the requested orange guitar. In (b), the outputs preserve reference-style characteristics such as the palette and illustrative treatment but omit the requested red pavilion and red clock tower, producing landscape scenes instead.

\section{Additional Ablation Studies}
\label{sec:supp_ablation}

\subsection{Component and Optimization Ablations}
\label{sec:supp_component_ablation}

We evaluate eight variants on ReConStyle-Full to examine auxiliary-layer conditioning, pathway complementarity, and optimization strategy. All variants use the same training data, backbone, and budget of 55,000 updates.

\paragraph{Variant Definitions.}
The three style-pathway variants use primary-layer conditioning at 1.0 and auxiliary scales of 0, 1.0, or 0.05. Native-only conditioning uses the native reference pathway without the style pathway. These four variants, corresponding to the first four rows of Table~\ref{tab:supp_component_ablation}, are trained for all 55,000 updates. The remaining variants start from a 40,000-update style pathway trained with the Stage-1 configuration in Appendix~\ref{sec:supp_prostyle_training} and differ only in the final 15,000 updates. Style pathway continuation retains weak auxiliary conditioning and continues training with a fresh optimizer at the Stage-3 rates in Table~\ref{tab:supp_training_configuration}. Joint optimization trains both pathways together throughout these updates without a frozen stage, whereas extended Stage 2 keeps the style pathway frozen during continued native-pathway training. ProStyle uses the three-stage schedule of prior learning, native residual fitting, and joint refinement.

\paragraph{Auxiliary-Layer Conditioning.}
Weak auxiliary conditioning raises Style-CSD from 0.3529 to 0.3811 while maintaining TCF at approximately 0.774. Compared with strong auxiliary conditioning, it achieves higher CLIP-T and TCF and lower ROO. These results support weak auxiliary conditioning as a way to extend style information beyond the primary layers while retaining target-content control.

\paragraph{Pathway Complementarity.}
Jointly optimized dual pathways raise Style-CSD and TCF to 0.3996 and 0.7756, compared with 0.3863 and 0.7551 for native-only conditioning, so the learned style prior adds to what full-reference features provide alone. Continuing only the style pathway at a low learning rate reaches 0.3731, so the dual-pathway gain does not come from additional training alone.

\paragraph{Progressive Optimization.}
Under the dual-pathway architecture, ProStyle achieves Style-CSD of 0.4076, compared with 0.3996 for joint optimization and 0.3998 for extended Stage 2, and also raises TCF from 0.7667 to 0.7729 and reduces ROO from 0.0318 to 0.0246 relative to the latter. These results support coordinating both pathways during the final refinement stage.

\begin{table}[t]
	\centering
	\caption{Effect of auxiliary conditioning strength on ReConStyle-Rapid. The primary-layer scale is fixed at 1.0. Bold indicates the best mean in each column.}
	\label{tab:supp_auxiliary_strength}
	\small
	\setlength{\tabcolsep}{3pt}
	\begin{tabular}{lcccc}
		\toprule
		Auxiliary scale & Style-CSD$\uparrow$ & CLIP-T$\uparrow$ & TCF$\uparrow$ & ROO$\downarrow$ \\
		\midrule
		0 & 0.3953 & 0.2388 & 0.7278 & 0.0967 \\
		0.025 & 0.3968 & \textbf{0.2394} & 0.7292 & 0.0979 \\
		0.05 & \textbf{0.4203} & 0.2323 & 0.6989 & \textbf{0.0932} \\
		0.1 & 0.4176 & 0.2393 & \textbf{0.7306} & 0.0952 \\
		\bottomrule
	\end{tabular}
\end{table}

\begin{table}[t]
	\centering
	\caption{Effect of auxiliary conditioning locations on ReConStyle-Rapid. The auxiliary scale is fixed at 0.05. Bold indicates the best mean in each column.}
	\label{tab:supp_auxiliary_locations}
	\small
	\setlength{\tabcolsep}{3pt}
	\begin{tabular}{lcccc}
		\toprule
		Auxiliary layers & Style-CSD$\uparrow$ & CLIP-T$\uparrow$ & TCF$\uparrow$ & ROO$\downarrow$ \\
		\midrule
		$\{4,5\}$ & 0.4055 & 0.2372 & 0.7244 & 0.0998 \\
		$\{6\}$ & 0.4070 & 0.2364 & 0.7078 & \textbf{0.0869} \\
		$\{7,8\}$ & 0.3970 & \textbf{0.2381} & 0.7228 & 0.1049 \\
		$\{4,5,6\}$ & 0.4151 & 0.2341 & 0.6969 & 0.1022 \\
		$\{6,7,8\}$ & 0.3985 & 0.2370 & \textbf{0.7297} & 0.0937 \\
		$\{4,5,6,7,8\}$ & \textbf{0.4203} & 0.2323 & 0.6989 & 0.0932 \\
		\bottomrule
	\end{tabular}
\end{table}
\subsection{Auxiliary-Layer Conditioning Strength and Placement}
\label{sec:supp_auxiliary_ablation}

We study how auxiliary-layer strength and placement affect the style pathway before adding native conditioning. Each variant uses 10,000 cross-content pairs from a 1,049-image subset spanning 15 styles and is trained for 5,000 updates. Evaluation uses the 300 conditions in ReConStyle-Rapid.

\paragraph{Conditioning Strength.}
With the primary-layer scale fixed at 1.0, we vary the auxiliary scale over $\{0,0.025,0.05,0.1\}$ during training and inference. In this short-schedule setting, Table~\ref{tab:supp_auxiliary_strength} shows that 0.05 achieves the highest Style-CSD and lowest ROO, with slightly lower TCF and CLIP-T than the other scales. Under the full training schedule in Appendix~\ref{sec:supp_component_ablation}, the same setting raises Style-CSD from 0.3529 to 0.3811 while TCF changes from 0.7733 to 0.7746, so we adopt it for ProStyle.

\paragraph{Conditioning Layers.}
With the auxiliary scale fixed at 0.05, we vary which auxiliary layers receive reference features. Primary layers retain scale 1.0, while unselected auxiliary layers use zero. Layer indices follow Appendix~\ref{sec:supp_prostyle_training}. Table~\ref{tab:supp_auxiliary_locations} shows that conditioning all five auxiliary layers achieves the highest Style-CSD, while different subsets favor the content metrics. We use all five auxiliary layers for their stronger style similarity.

\begin{table}[t]
	\centering
	\caption{Effect of the number of style tokens on ReConStyle-Rapid. All variants use $\lambda_n=1$. Bold indicates the best mean in each column.}
	\label{tab:supp_tokens}
	\small
	\setlength{\tabcolsep}{3pt}
	\begin{tabular}{lcccc}
		\toprule
		Tokens & Style-CSD$\uparrow$ & CLIP-T$\uparrow$ & TCF$\uparrow$ & ROO$\downarrow$ \\
		\midrule
		8 & 0.4011 & \textbf{0.2414} & 0.7608 & \textbf{0.0859} \\
		16 & \textbf{0.4295} & 0.2410 & 0.7578 & 0.0874 \\
		32 & 0.4291 & 0.2377 & \textbf{0.7628} & 0.0874 \\
		\bottomrule
	\end{tabular}
\end{table}

\begin{table}[t]
	\centering
	\caption{Effect of the native pathway weight $\lambda_n$ on ReConStyle-Rapid with 16 style tokens. Bold indicates the best mean in each column.}
	\label{tab:supp_native_weight}
	\small
	\setlength{\tabcolsep}{3pt}
	\begin{tabular}{lcccc}
		\toprule
		$\lambda_n$ & Style-CSD$\uparrow$ & CLIP-T$\uparrow$ & TCF$\uparrow$ & ROO$\downarrow$ \\
		\midrule
		0.75 & 0.4291 & 0.2405 & \textbf{0.7650} & \textbf{0.0868} \\
		1.0 & \textbf{0.4295} & \textbf{0.2410} & 0.7578 & 0.0874 \\
		1.25 & 0.4264 & 0.2405 & 0.7517 & 0.0869 \\
		\bottomrule
	\end{tabular}
\end{table}

\subsection{Number of Style Tokens}
\label{sec:supp_token_ablation}

We vary the number of style tokens on the training subset of Appendix~\ref{sec:supp_auxiliary_ablation}, using the full progressive schedule of 5,000 style-pathway updates, 500 native-only updates, and 4,500 joint updates, and evaluate on ReConStyle-Rapid. Table~\ref{tab:supp_tokens} shows that reducing the tokens from 16 to 8 lowers Style-CSD from 0.4295 to 0.4011, suggesting that eight tokens are too few to retain the stylistic information of the reference. Increasing the tokens to 32 brings no further gain in Style-CSD, which reaches 0.4291, and lowers CLIP-T from 0.2410 to 0.2377. We therefore use 16 tokens.

\subsection{Native Pathway Weight}
\label{sec:supp_native_weight}

We keep the training subset, progressive schedule, and ReConStyle-Rapid evaluation of Appendix~\ref{sec:supp_token_ablation} with 16 style tokens, and vary only the native pathway weight $\lambda_n$ in Eq.~\eqref{eq:dual_pathway}. Table~\ref{tab:supp_native_weight} shows that Style-CSD stays between 0.4264 and 0.4295 and TCF between 0.7517 and 0.7650 as $\lambda_n$ ranges from 0.75 to 1.25, and $\lambda_n=1$ gives the highest Style-CSD and CLIP-T.

\subsection{Self-Paired versus Cross-Content Training}
\label{sec:supp_self_pair_ablation}

This control tests the role of content differences between reference and target. We replace the reference with its target in every stage, yielding $(I_t,I_t,c_t)$ instead of cross-content tuples $(I_r,I_t,c_t)$. Cross-content tuples share style but differ in subject and background. Both variants use the full architecture, the same 134,000 target-sample entries and content prompts, and the same number of training updates. Target order and multiplicity remain fixed, with only reference assignment changed.

Table~\ref{tab:cross_content_pairing} evaluates both strategies on all 50 Full styles and 6,600 conditions.

Compared with identity self-pairing, cross-content training lowers ROO and improves CLIP-T and TCF, while yielding lower category-level Style-CSD. These results support the role of cross-content supervision in preserving target content and suppressing reference-content leakage.

\begin{table}[t]
	\centering
	\caption{Training data and architecture on ReConStyle-Full. All variants except the released DEADiff are trained on ReConStyle. Bold indicates the best value in each column.}
	\label{tab:supp_data_architecture}
	\small
	\setlength{\tabcolsep}{3pt}
	\begin{tabular}{lcccc}
		\toprule
		Method & Style-CSD$\uparrow$ & CLIP-T$\uparrow$ & TCF$\uparrow$ & ROO$\downarrow$ \\
		\midrule
		DEADiff (released) & 0.3750 & 0.2565 & 0.6596 & 0.0290 \\
		DEADiff (retrained) & 0.3492 & \textbf{0.2682} & 0.7141 & \textbf{0.0215} \\
		Style pathway only & 0.3529 & 0.2657 & \textbf{0.7733} & \textbf{0.0215} \\
		ProStyle & \textbf{0.4076} & 0.2668 & 0.7729 & 0.0246 \\
		\bottomrule
	\end{tabular}
\end{table}

To separate the effects of training data and architecture, we retrain the style branch of DEADiff on the 134,000 ReConStyle pairs from SD1.5 and BLIP-Diffusion initialization, with 55,000 updates at batch size 8, and evaluate it with the inference settings of Table~\ref{tab:main_comparison}. Table~\ref{tab:supp_data_architecture} shows that retraining raises its TCF from 0.6596 to 0.7141 and CLIP-T from 0.2565 to 0.2682 and lowers ROO from 0.0290 to 0.0215. With the same data, ProStyle reaches Style-CSD and TCF of 0.4076 and 0.7729, compared with 0.3492 and 0.7141 for the retrained DEADiff.

\end{document}